\documentclass[10pt,twocolumn,letterpaper]{article}
\usepackage[accsupp]{axessibility}  
\usepackage{iccv}              
\usepackage{wrapfig}
\usepackage{multirow} 
\usepackage{amssymb}
\usepackage{pifont}
\newcommand{\cmark}{\ding{51}}%
\usepackage{algorithm}
\usepackage{algorithmic}
\usepackage{bbm}
\usepackage{graphicx}
\usepackage{enumitem}
\usepackage{booktabs}
\usepackage{colortbl}
\usepackage{stfloats}

\usepackage{times}

\definecolor{iccvblue}{rgb}{0.21,0.49,0.74}
\usepackage[pagebackref,breaklinks,colorlinks,allcolors=iccvblue]{hyperref}

\def\paperID{1332} 
\def\confName{ICCV}
\def\confYear{2025}

\title{Cooperative Pseudo Labeling for Unsupervised Federated Classification}

\author{Kuangpu Guo$^{1,2}$, Lijun Sheng$^{1,2}$, Yongcan Yu$^{2,3}$, Jian Liang\thanks{Corresponding author} \ $^{2,3}$, Zilei Wang$^1$ , Ran He$^{2,3}$\\
$^1$ University of Science and Technology of China\\
$^2$ NLPR \& MAIS, Institute of Automation, Chinese Academy of Sciences\\
$^3$ School of Artificial Intelligence, University of Chinese Academy of Sciences \\
{\tt\small gkp@mail.ustc.edu.cn, liangjian92@gmail.com}
}

\begin{document}
\maketitle
\begin{abstract}
Unsupervised Federated Learning (UFL) aims to collaboratively train a global model across distributed clients without sharing data or accessing label information.
Previous UFL works have predominantly focused on representation learning and clustering tasks.
Recently, vision language models (e.g., CLIP) have gained significant attention for their powerful zero-shot prediction capabilities.
Leveraging this advancement, classification problems that were previously infeasible under the UFL paradigm now present promising new opportunities, yet remain largely unexplored.
In this paper, we extend UFL to the classification problem with CLIP for the first time and propose a novel method, \underline{\textbf{Fed}}erated \underline{\textbf{Co}}operative \underline{\textbf{P}}seudo \underline{\textbf{L}}abeling (\textbf{FedCoPL}). 
Specifically, clients estimate and upload their pseudo label distribution, and the server adjusts and redistributes them to avoid global imbalance among classes.
Moreover, we introduce a partial prompt aggregation protocol for effective collaboration and personalization.
In particular, visual prompts containing general image features are aggregated at the server, while text prompts encoding personalized knowledge are retained locally.
Extensive experiments demonstrate the superior performance of our FedCoPL compared to baseline methods.
Our code is available at \href{https://github.com/krumpguo/FedCoPL}{https://github.com/krumpguo/FedCoPL}.
\end{abstract}
 
\section{Introduction}
\label{sec:intro}
Federated learning (FL) \cite{sheller2020federated, li2020federated} is a distributed machine learning paradigm that facilitates collaboration among clients while preserving data privacy by avoiding data sharing~\cite{li2022federated, shi2023understanding}.
Conventional FL predominantly follows a supervised learning paradigm~\cite{zhang2025personalized, guo2024not, caldarola2024beyond, Xinting2025FOCoOp, guo2024dynamic}, relying on well-annotated labels at the client side. 
However, data annotation is often time-consuming and requires domain expertise~\cite{luo2021no, zhang2023fedala,rehman2023dawa,shi2023towards}, posing challenges in real-world applications.
Consequently, substantial research efforts have been dedicated to unsupervised federated learning (UFL)~\cite{tiantowards, kim2023protofl, han2022fedx, yan2024lightweight}, which facilitates collaborative model training among clients using unlabeled data.
Existing UFL works~\cite{zhang2023federated, pourpanah2024federated, liao2024rethinking} primarily focus on leveraging unlabeled data for representation learning or clustering, enabling models to extract features that enhance generalization and performance on downstream tasks.
However, the absence of labeled data prevents the direct application of UFL to classification tasks.

Vision-language models (VLMs)~\cite{li2019visualbert, radford2021learning, alayrac2022flamingo}, particularly contrastive language-image pre-training (CLIP)~\cite{radford2021learning}, have recently garnered significant attention for their remarkable representation and generalization capabilities~\cite{wen2023improving, faysse2024colpali, marcu2024lingoqa}. 
By pre-training on large-scale image-text pairs, CLIP extracts modality-specific features through two encoders, enabling powerful zero-shot classification. 
The impressive zero-shot prediction capability of CLIP alleviates the dependency on labeled data.
This advancement presents new opportunities for addressing previously intractable classification tasks in UFL, yet remains largely unexplored.

In this paper,  we are the first to extend UFL to classification tasks by leveraging CLIP.
Each client is initialized with the same pre-trained CLIP model and possesses only unlabeled local data.
Similar to standard federated learning frameworks~\cite{xie2024perada, yang2024fedas}, clients can periodically upload their non-data knowledge (e.g., statistical information, model parameters) to the central server for aggregation and receive updated global knowledge in return.
However, when directly applied within the UFL framework, the inherent bias of CLIP and the unknown label distribution among clients~\cite{huang2022unsupervised, menghini2023enhancing, luo2021no} can result in globally imbalanced training across categories.
Moreover, the potential label skew can introduce discrepancies in model update directions among clients, rendering conventional aggregation methods ineffective for collaboration among clients.

To address these challenges, we propose a novel method, \textbf{FedCoPL}, which comprises two key components: cooperative pseudo labeling and partial prompt aggregation.
The cooperative pseudo labeling mechanism aims to construct a more representative pseudo label distribution while ensuring balanced training across all classes.
At the beginning of federated training, each client leverages confidence-based and entropy-based filtering methods to estimate its pseudo label distribution, which is then transmitted to the server.
The server globally adjusts the distribution for each client based on the proportion of pseudo labels in each category.
This process alleviates class bias caused by CLIP’s inherent bias and the unknown data distribution among clients, ultimately enhancing the quality of pseudo labels.

To promote effective collaboration and personalization, we introduce a partial prompt aggregation strategy.
Specifically, for efficient computation and communication, we only optimize the visual and textual prompts instead of the entire model parameters.
Previous studies~\cite{xing2023dual} have shown that visual prompts tend to capture general representations of the feature space, whereas textual prompts are more focused on category-specific information. 
Aggregating prompts from both modalities simultaneously may lead to parameter conflicts, especially in the presence of label skew.
Therefore, we propose transmitting only visual prompts to the server for aggregation, while retaining textual prompts locally.
This strategy enhances global collaboration through aggregated visual prompts while enabling clients to benefit from personalized textual prompts, which better align with their local label distributions.
Experimental results on standard federated prompt learning benchmarks, with both Dirichlet-based and quantity-based label skews, demonstrate the effectiveness of the proposed FedCoPL compared to baseline methods.
Our main contributions are summarized as follows:
\begin{itemize}

\item This paper extends the unsupervised federated learning paradigm to classification tasks with CLIP, which remains largely unexplored in previous works.
\item We propose FedCoPL to mitigate label skew and the inherent bias of CLIP. FedCoPL primarily incorporates a cooperative pseudo labeling strategy to construct more representative and accurate pseudo labels.
\item FedCoPL further integrates a partial prompt aggregation protocol to facilitate effective collaboration while preserving client personalization.
\item Extensive experiments across six commonly used datasets with Dirichlet-based and quantity-based label skews demonstrate the effectiveness of FedCoPL.
\end{itemize}

\section{Related Work}
\label{2_related_work}
\subsection{Unsupervised Federated Learning}
Unsupervised federated learning aims to collaboratively train a global model across distributed clients with unlabeled data while preserving data privacy by avoiding data sharing. 
Due to the lack of prior label knowledge, existing unsupervised federated methods primarily concentrate on representation learning and clustering tasks.
Specifically, both FedU~\cite{zhuang2021collaborative} and FedEMA~\cite{zhuang2022divergence} enhance the awareness of heterogeneity in unsupervised federated learning by divergence-aware predictor update rule, and adaptive global knowledge interpolation, respectively.
Additionally, FedUL~\cite{lu2022federated} introduces federated learning with only unlabeled data but requires knowledge of precise label frequencies for each client. 
L-DAWA~\cite{rehman2023dawa} proposes prototypical representation distillation to learn a representation from unlabeled data.
FedUU~\cite{liao2024rethinking} uses flexible uniform regularize to obtain uniform and unified representation and FedGaLA~\cite{pourpanah2024federated} performs gradient alignment at the client level to encourage clients to learn domain-invariant features.
Moreover, Orchestra~\cite{lubana2022orchestra} utilizes local-global clustering to guide self-supervised learning. 
In contrast, we propose leveraging CLIP's zero-shot capability to generate pseudo labels for model training, enabling the handling of more complex tasks like image classification.

\subsection{Federated Learning for VLMs}
The fine-tuning of vision-language models recently has been extended to the federated framework to alleviate the computational load while addressing challenges in federated learning, such as robustness in cross-domain scenarios and non-IID data distributions~\cite{qiu2023text, li2024position, halbe2023hepco, su2022cross}.
For example, FedCLIP~\cite{lu2023fedclip} and PromptFL~\cite{guo2023promptfl} directly extend the standard fine-tuning of CLIP to the federated setting to achieve strong performance.
pFedprompt~\cite{guo2023pfedprompt} combines a federated prompt learning scheme with personalized spatial visual features.
Additionally, pFedPG~\cite{yang2023efficient} generates personalized prompts for each client based on their visual prompts to better align with their data distribution.
FedOPT~\cite{li2024global} utilizes knowledge from both personal and global textual prompts for prediction through unbalanced optimal transport.
pFedMoAP~\cite{luo2024mixture} personalizes the prompt learning process through the lens of Mixture of Experts.
However, previous methods have primarily focused on supervised federated learning with VLMs. 
In this paper, we explore leveraging VLMs for unsupervised federated classification, capitalizing on their zero-shot capabilities.

\begin{figure*}[tbp]
    \centering
    \includegraphics[width=0.95\textwidth]{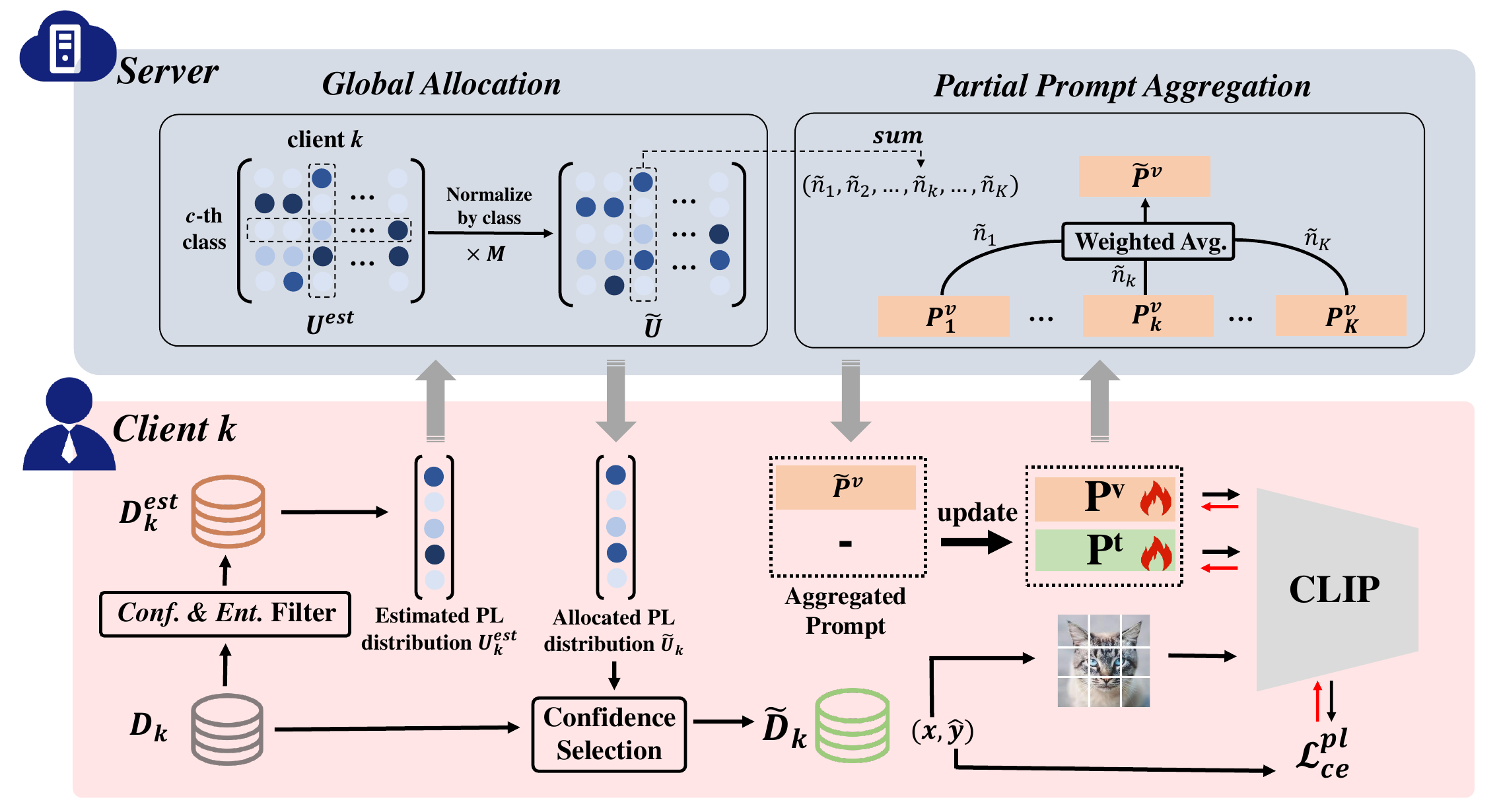}
    \caption{
    \textbf{The overview of our FedCoPL.} For pseudo labeling, FedCoPL begins by filtering the initial unlabeled samples to estimate the local distributions, which are uploaded to the server. Then, the server globally selects $M$ pseudo labels for each category and allocates them to clients based on local estimated distributions.
    To handle label skews during training, we aggregate only visual prompts on the server to enhance global performance because the differences in textual prompts are significantly greater than those found in visual prompts.} 
    \label{fig:method-client}
\end{figure*}

\subsection{Unsupervised Learning for VLMs}
In real-world applications, high annotation costs are often required to ensure that each data source has labeled data. 
This necessity drives us to develop effective methods for utilizing unlabeled data in downstream tasks.
Pseudo labeling strategy~\cite{huang2022unsupervised, menghini2023enhancing, zhang2024candidate, jia2024mltu, tanwisuth2023pouf} and entropy minimization~\cite{liangrealistic} have been widely studied in this context.
For instance, UPL~\cite{huang2022unsupervised} and FPL~\cite{menghini2023enhancing} select an equal number of pseudo labels for each category, while CPL~\cite{zhang2024candidate} generates multiple candidate labels per instance to improve labeling accuracy.
POUF~\cite{tanwisuth2023pouf} fine-tunes the model or prompt through the alignment of discrete label distributions between the source prompts and the unlabeled target data.
UEO~\cite{liangrealistic} adopts a confidence-aware strategy, minimizing the conditional entropy of high-confidence samples while simultaneously maximizing the marginal entropy of low-confidence ones
Despite their effectiveness, these pseudo labeling strategies face substantial challenges when directly applied to decentralized unsupervised federated settings~\cite{huang2022unsupervised}.
To address this challenge, we propose a novel cooperative pseudo-labeling strategy that mitigates the inherent biases of pre-trained VLMs and significantly improves the quality of pseudo labels.

\section{Method}
In this section, we provide a detailed formulation of unsupervised federated learning for classification tasks with CLIP and introduce our proposed method.
First, in Sec.~\ref{Preliminaries}, we review key principles underlying CLIP~\cite{radford2021learning} and prompt tuning.
Next, Sec.~\ref{Unsupervised Heterogeneous Federated Framework for VLMs} outlines the framework of unsupervised federated learning with CLIP and presents its associated challenges.
We then introduce our proposed method, FedCoPL, which is built upon two key strategies: the cooperative pseudo labeling strategy discussed in Sec.~\ref{Federated Adaptive Pseudo-label Selection}, and the partial prompt aggregation protocol covered in Sec.~\ref{Partial Prompt Aggregation}.

\subsection{Preliminary}
\label{Preliminaries}
We employ CLIP~\cite{radford2021learning} as the foundational model for our framework.
CLIP utilizes a dual-branch architecture consisting of an image encoder, $f^v(\cdot)$, and a text encoder, $f^t(\cdot)$, where each encoder processes data from its respective modality.
For zero-shot predictions in downstream tasks, CLIP utilizes a human-designed prompt (e.g., “a photo of a [CLASS]”) for each class.
Consider a C-way classification task as an example, textual embeddings of all classes $\{f^t_c\}_{c=1}^C$ and the visual embedding of the test image $f^v(x)$ are derived from the text and image encoders, respectively.
The probability that image $x$ belongs to the $c$-th category is calculated by applying the softmax operation as follows:
\begin{equation}
p_c(x) = \frac{\exp(\text{sim}(f^v(x), f^t_c)) / \tau)}{\sum_{j=1}^{C} \exp(\text{sim}(f^v(x), f^t_j)) / \tau)},
\end{equation}
where $\tau$ is a temperature parameter.
To enhance performance in downstream tasks, prompt tuning is widely adopted as a parameter-efficient fine-tuning method~\cite{zhou2022learning}.
This involves introducing additional learnable textual tokens $P^t$ and visual tokens $P^v$ ~\cite{jia2022visual, xing2023dual} (referred to as textual and visual prompts) into the corresponding encoders, to adapt the original CLIP model to specific downstream tasks.

\subsection{Unsupervised Federated Learning with CLIP}
\label{Unsupervised Heterogeneous Federated Framework for VLMs}
The powerful zero-shot classification capability of CLIP reduces reliance on labeled data, enabling the solution of previously intractable classification tasks in the context of UFL.
Therefore, we extend the UFL paradigm to classification tasks by leveraging CLIP.
In this setting, each client is initialized with the same pre-trained CLIP model and possesses an unlabeled local dataset ${D_k}$ with a capacity of ${n_k}$.
Additionally, each client is provided with the full set of class names for the classification task.
Similar to standard federated learning frameworks, clients periodically upload non-data knowledge (e.g., statistical information, model parameters) to the server for aggregation and download global updates in return.
To avoid ambiguity, we clarify that the label distribution and label skew referenced in this paper refer to the ground-truth labels of clients' datasets, which remain unknown during training.

The primary challenge arises from the inherent bias of CLIP and the data heterogeneity, particularly the label skew across clients.
Specifically, while the union of all client datasets $\bigcup_{k \in K} D_k$ encompasses samples from all categories, each individual dataset $D_k$ may contain only a subset of these samples.
Therefore, the direct application of CLIP within the UFL framework can result in an imbalanced training process across different categories.
Additionally, the label skew among clients can lead to discrepancies in the model update directions across clients, which can hinder effective collaboration.

\subsection{Cooperative Pseudo Labeling}
\label{Federated Adaptive Pseudo-label Selection}
To address the complex challenges inherent in unsupervised heterogeneous federated learning, we propose an effective and robust framework, FedCoPL.
Our method incorporates two core strategies: a cooperative pseudo labeling (CoPL) strategy and a partial prompt aggregation protocol.
A visual overview of the framework is provided in Fig.~\ref{fig:method-client}, with detailed pseudocode presented in the supplemental material.

Given the inherent bias of CLIP and the presence of label skews in the unlabeled data across clients, the direct application of conventional pseudo labeling methods can result in globally unbalanced and weak representative pseudo labels.
Specifically, conventional pseudo labeling methods tend to select pseudo labels from categories favored by CLIP, even when these pseudo labels are incorrect~\cite{menghini2023enhancing,zhang2024candidate}.
To address this, we introduce a cooperative pseudo labeling strategy to select a more representative pseudo label distribution and ensure a balanced distribution across all classes.

Specifically, for each client $k$, we filter samples from the original unlabeled set $D_k$ according to model prediction confidence and entropy~\cite{huang2022unsupervised, shu2022test} to construct the estimated set $D^{est}_k$ as follows:
\begin{equation}
\label{eq2}
\hspace*{-0.05cm}
D^{est}_k = \{(x, \hat{y}) |\max\limits_{c} p_c(x) > \tau _1 \ , \ Ent(p(x)) < \tau _2 \},
\end{equation}
where $\hat{y} = \arg\max\limits_{c} p_c(x)$ represents the predicted pseudo label, and $Ent(\cdot)$ denotes the Shannon Entropy operator.
Here, $\tau_1$ and $\tau_2$ represent the medians of the pseudo label confidences and the entropies of the training set, respectively.
Due to the bias in CLIP, $D^{est}_k$ tends to contain pseudo labels from categories favored by CLIP, even when these pseudo labels are incorrect~\cite{menghini2023enhancing,zhang2024candidate}.
For example, suppose the true label distribution of $D_k$ is [0.1, 0.1, 0.8], while the pseudo label distribution of $D^{est}_k$ 
maybe [0.6, 0.05, 0.35], which has low accuracy and fails to represent the true label distribution.
To address this issue, we propose a global allocation strategy that redistributes adjusted pseudo label distributions to clients in proportion to their locally estimated label distributions. 

To make clients’ pseudo label distributions more representative, we compute category-wise statistics on the filtered pseudo labels, yielding the estimated distribution $U^{est}_k$, a $C$-dimensional vector where the $c$-th element $u_{k,c} = \sum_{i} \mathbbm{1} (\hat{y}_i = c)$ indicates the number of samples associated with class $c$ in client $k$ according to the pseudo labels.
Then, clients upload their estimated distribution $U^{est}_k$ to the server for the collaborative allocation.
Following~\cite{ye2023feddisco, zhang2024candidate}, we assume a uniform joint label distribution across all clients.
To ensure adequate training for all classes, the server globally selects $M$ pseudo labels for each category and allocates them to clients based on estimated distributions, as follows:
\begin{align} 
\label{eq:U_def}
\widetilde{U}_k &= (\widetilde{u}_{k,1}, \widetilde{u}_{k,2}, ..., \widetilde{u}_{k,C}), \nonumber \\
\widetilde{u}_{k,c} &= \left\lceil \frac{u_{k,c}}{\sum_{i \in K}{u_{i,c}}} \cdot M \right\rceil,
\end{align}

where $M= \frac {\sum_{k \in K}\sum_{c \in C}{u_{k,c}}}{C} $ is the global number of pseudo labels per class and $\widetilde{u}_{k,c}$ denotes the amount of pseudo labels of category $c$ assigned to client $k$.

Finally, client $k$ constructs its training set $\widetilde{D}_k$ based on the adjusted capacity $\widetilde{U}_{k}$.
Specifically, for each class $c$, the $\widetilde{u}_{k,c}$ samples with the highest prediction probabilities in the original dataset $D_k$ are selected and incorporated into $\widetilde{D}_k$.
As described in Eq.~(\ref{eq:U_def}), the overall sample budget $M$ is allocated proportionally according to the clients’ estimated distributions within each category, rather than relying on the absolute values of these estimates.
This allocation strategy guarantees that, even when the estimated distributions are noisy or fail to fully capture the client’s true local label distribution, the pseudo labels selected by our method remain both representative and accurate, thereby preserving the essential characteristics of the client’s data and supporting robust collaborative learning.

As federated training progresses, we will periodically repeat these steps to generate, estimate, assign, and select pseudo labels.
This iterative procedure ensures that the pseudo-labeled training data $\widetilde{D}_k$ is regularly updated, maintaining a globally balanced and high-accuracy dataset.

\subsection{Partial Prompt Aggregation}
\label{Partial Prompt Aggregation}

\begin{figure}
    \small
    \setlength\tabcolsep{1mm}
    \renewcommand\arraystretch{0.1}
    \begin{tabular}{cc}
        \includegraphics[width=0.44\linewidth]{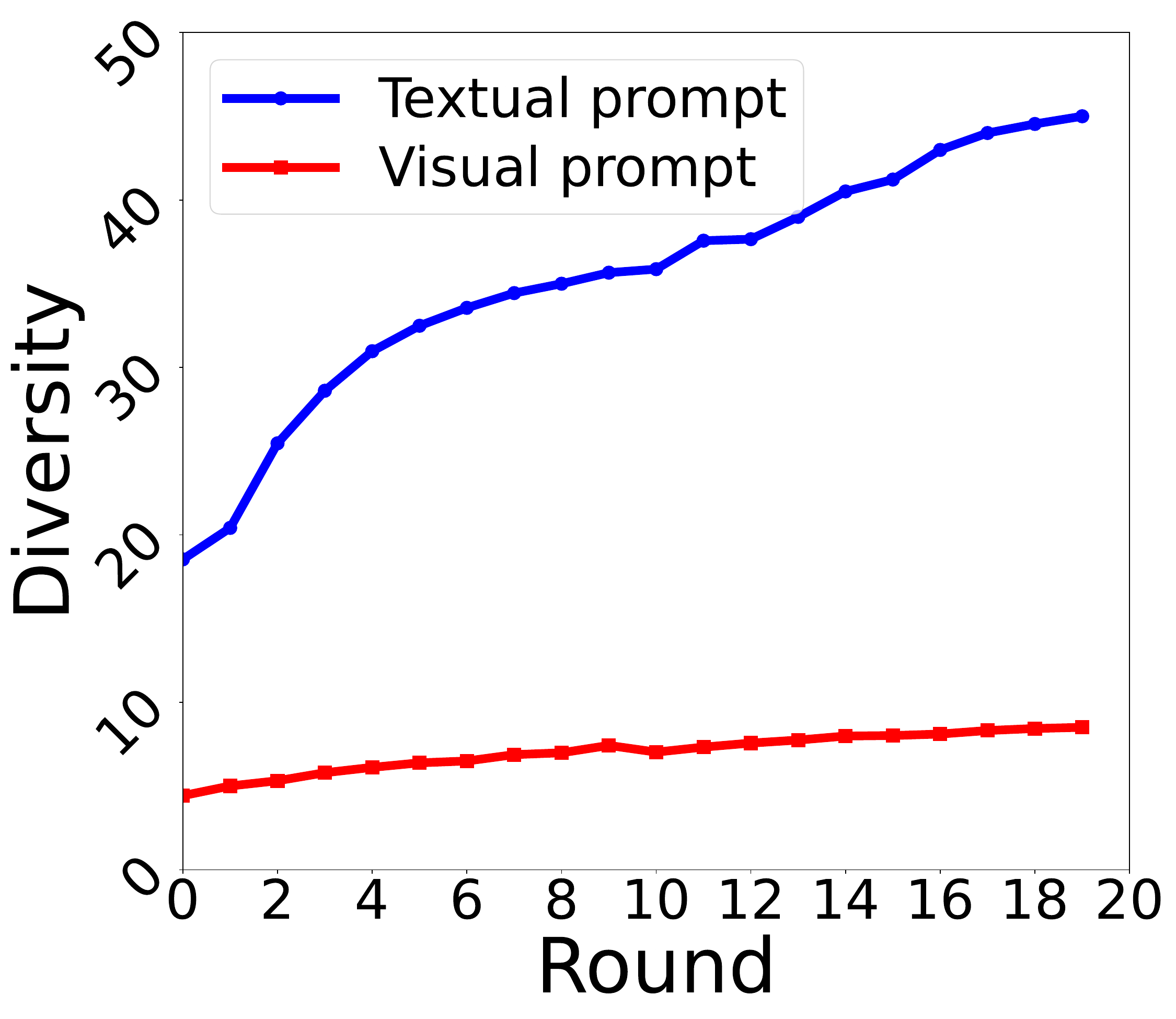} &		
        \includegraphics[width=0.44\linewidth, clip]{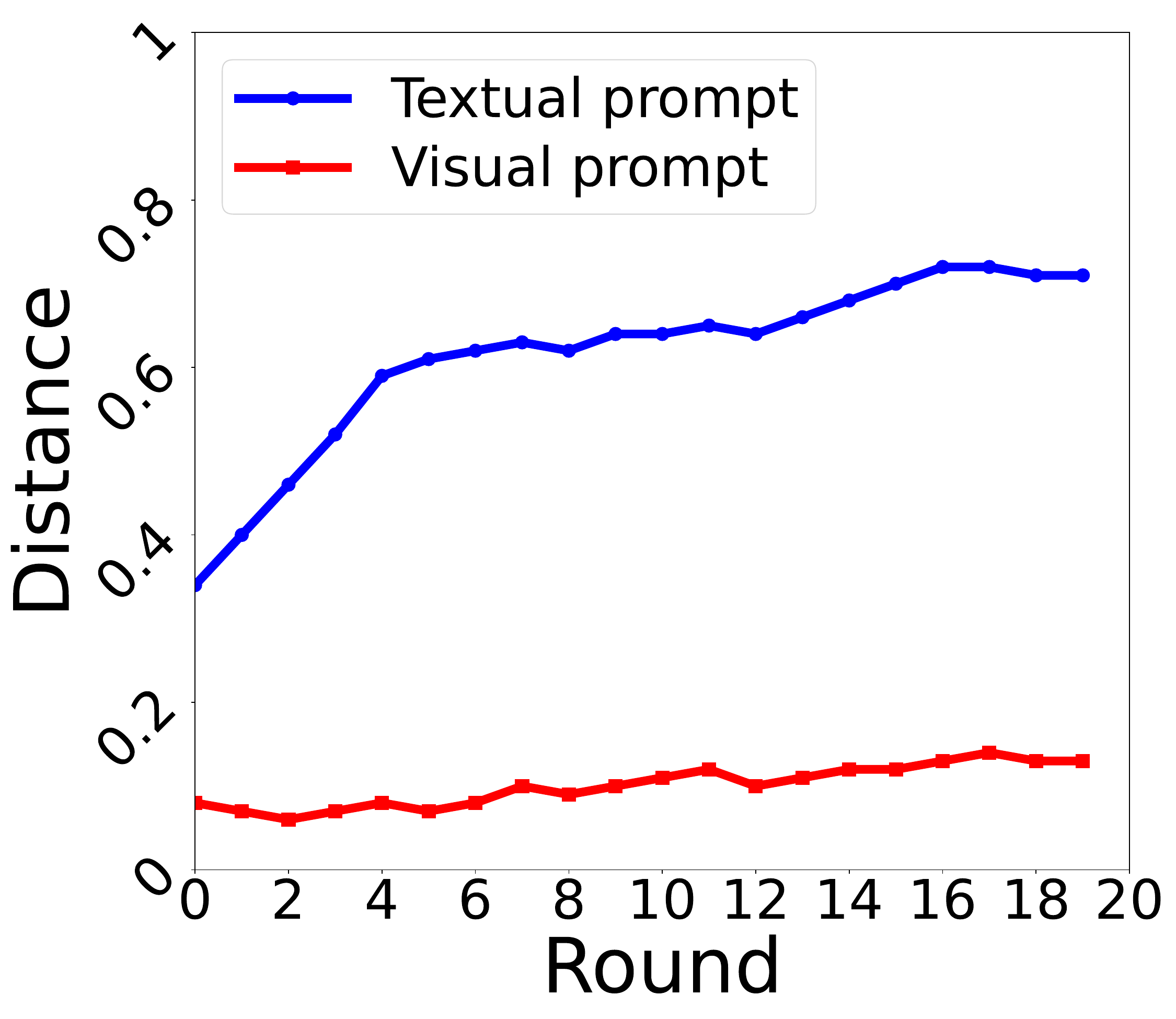} 
    \\[1.0mm]
        (a) Drift Diversity. & (b) Cosine Distance.
    \end{tabular}
    \caption{Drift diversity and cosine distance of prompts among clients during training in CIFAR10~\cite{krizhevsky2009learning} dataset.
    The differences observed in textual prompts are significantly greater than those observed in visual prompts.}
    \label{fig:shift}
\end{figure}

For each client, we select both textual prompt $P^t_k$ and visual prompt $P^v_k$ to be optimized using the local training set $\widetilde{D}_k$.
Since each client's training set pertains to the same task, effective knowledge aggregation can enhance the overall framework's performance.
A straightforward approach is to aggregate the prompts using a simple averaging operation.
However, due to label skews among clients, this may lead to suboptimal or even detrimental performance.

As for CLIP, updates to the visual branch primarily enhance image representation knowledge, while the textual branch focuses on determining the classification boundaries by leveraging category information~\cite{xing2023dual}.
Consequently, under heterogeneous data distributions among clients, where clients share the same task but different label distributions, the visual prompts $\{P^v_k\}_{k=1}^K$ are more likely to be similar, while the textual prompts $\{P^t_k\}_{k=1}^K$ exhibit greater variability.
To validate this conjecture, we measure the differences in both visual and textual prompts across all clients using drift diversity~\cite{li2023effectiveness} and cosine distance, which capture the magnitude and directional differences in prompt updates, respectively.
As illustrated in Fig.~\ref{fig:shift}, the differences in textual prompts are significantly greater than those in visual prompts, which confirms our hypothesis.

To avoid conflicts when aggregating widely different textual prompts and to promote cooperation among clients, we propose a partial prompt aggregation protocol. 
In this protocol, all clients upload their visual prompts $\{P^v_k\}_{k=1}^K$ to the server for aggregation while keeping their textual prompts $\{P^t_k\}_{k=1}^K$ locally for personalization.
For the server, we design the visual prompt aggregation strategy that utilizes a weighted average approach following~\cite{mcmahan2017communication} as follows:
\begin{equation}
\label{aggregate visual prompt}
{P}^v = \sum\limits_{k=1}^K \frac{\widetilde{n}_k}{\sum\nolimits_{i \in K}  \widetilde{n}_i} \cdot {P}^v_k,
\end{equation}
where $\widetilde{n}_k = \sum\limits_{c} \widetilde{u}_{k,c}$ represents the total number of samples assigned to client $k$.
The aggregated visual prompts ${P}^v$ are then distributed back to the clients as initialization for their local models for the next training round.
The overall objective function of FedCoPL is formulated as follows:
\begin{equation}
\min\limits_{{P}^v, \{P^t_k\}^{K}_{k=1}} \sum\limits_{k=1}^{K}\mathbb{E}_{(x,\hat{y}) \in \widetilde{D}_k} {\ell}_{ce} \left( g({P}^v, P^t_k; x), \hat{y} \right),
\end{equation}
where $g({P}^v, P^t_k; \cdot)$ is the model output, $\hat{y}$ denotes pseudo labels, and ${\ell}_{ce}(\cdot, \cdot)$ is the cross-entropy loss function.

This approach enhances global performance by aggregating visual prompts while allowing clients to utilize personalized textual prompts that align better with their specific data distributions. 
Moreover, in comparison to methods that aggregate prompts from both modalities, our approach can reduce communication overhead by aggregating only textual prompts, making it more practical for environments where communication resources are constrained.

\textbf{Remark}. 
The proposed FedCoPL framework preserves client privacy by ensuring that the uploaded estimated distributions are derived solely from CLIP’s predictions, making it infeasible to inversely infer the clients’ exact label distributions. 
This highlights the enhanced privacy-preserving nature of our method, especially when compared with several existing approaches~\cite{zhang2022fine, kim2022federated} that require the transmission of exact category distributions, potentially exposing sensitive client-specific information.

\begin{table*}[t]
\setlength{\tabcolsep}{2pt}
\centering
\caption{Accuracy (\%) of different methods under Dirichlet-based label skews with ViT/B-32 as the backbone. FPL~\cite{menghini2023enhancing} and CPL~\cite{zhang2024candidate} are adopted as baseline pseudo labeling (\textbf{PL}) method. \textbf{Bold} and \underline{underline} represent the best and second-best results.}
\label{tab: accuracy Dirichlet}
\resizebox{\textwidth}{!}{ 
\begin{tabular}{lccccccccccccccc}
\toprule 
\multirow{2}{*}{Method} & \multirow{2}{*}{\textbf{PL}} & \multicolumn{2}{c}{{DTD}} & \multicolumn{2}{c}{{RESISC45}} & \multicolumn{2}{c}{{CUB}} & \multicolumn{2}{c}{{UCF101}} & \multicolumn{2}{c}{{CIFAR10}} & \multicolumn{2}{c}{{CIFAR100}} & \multicolumn{2}{c}{{Average}} \\ \cmidrule{3-16} 
    & & $\beta = 0.1$ & $\beta = 0.05$ & $\beta = 0.1$ & $\beta = 0.05$& $\beta = 0.1$ & $\beta = 0.05$& $\beta = 0.1$ & $\beta = 0.05$& $\beta = 0.1$ & $\beta = 0.05$ & $\beta = 0.1$ & $\beta = 0.05$  & $\beta = 0.1$ & $\beta = 0.05 $\\ \midrule
CLIP & - & {43.24} &{43.24} & {54.51} & {54.51} & \underline{51.28} & \underline{51.28} & {61.00} & {61.00} & {86.93} &  {86.93} & {64.17} & {64.17} & 60.19 & 60.19 \\ 	
PromptFL~\cite{guo2023promptfl} & FPL & 45.79 & 44.62 & 59.76 & 58.05 & 47.29 & 46.16 & 64.39 & 62.96 & 87.51 & 85.02 & 63.26 & 62.56 & 61.33 & 59.90 \\	
PromptFL~\cite{guo2023promptfl} & CPL & 44.84 & 46.32 & 62.52 & 60.99 & 48.72 & 48.86 & 63.86 & 64.97 & 88.90 & 89.14 & 66.04 & 65.14 & 62.48 & 62.57 \\
Promptprox~\cite{li2020federated} & FPL & 45.15 & 43.88 & 59.36 & 58.59 & 47.04 & 47.01 & 62.91 & 61.27 & 87.13 & 85.62 & 62.77 & 64.12 & 60.73 & 60.08 \\
Promptprox~\cite{li2020federated} & CPL & 43.51 & 45.85 & 59.99 & 62.35 & 49.25 & 48.65 & 64.55 & 63.94 & 90.44 & 90.07 & 65.86 & 66.15 & 62.27 & 62.84 \\	
pFedPrompt~\cite{guo2023pfedprompt} & FPL & 44.56 & 46.19 & \underline{65.95} & 60.52 & 48.48 & 44.42 & 64.37 & 64.18 & 87.45 & 90.72 & 65.08 & 65.83 & 62.65 & 61.98 \\	
pFedPrompt~\cite{guo2023pfedprompt} & CPL & 44.22 & 47.59 & 61.76 & \underline{66.79} & 47.23 & 50.97 & \underline{65.59} & \underline{65.44} & 90.26 & 89.10 & 65.63 & \underline{67.86} & 62.45 & \underline{64.63} \\	
FedOPT~\cite{li2024global} & FPL & \underline{51.97} & \underline{48.71} & 57.76 & 55.62 & 47.99 & 47.05 & 61.83 & 64.56 & \underline{91.32} & 90.73 & \underline{67.51} & 66.50 & \underline{63.06} & 62.20 \\	
FedOPT~\cite{li2024global} & CPL & 37.43 & 39.33 & 51.10 & 48.47 & 46.61 & 45.07 & 58.93 & 57.12 & 91.08 & \underline{91.36} & 57.87 & 59.81 & 57.17 & 56.86 \\\cmidrule{1-16} \rowcolor[gray]{0.9}
\textbf{FedCoPL} & \textbf{CoPL} & \textbf{60.89} & \textbf{66.37} & \textbf{75.76} & \textbf{80.26} & \textbf{56.09} & \textbf{54.80} & \textbf{73.20} & \textbf{74.97} & \textbf{95.38} & \textbf{95.11} & \textbf{73.59} & \textbf{72.84}  & \textbf{72.49} & \textbf{74.06} \\
\bottomrule
\end{tabular}
}
\end{table*}

\begin{table*}[t]
\setlength{\tabcolsep}{8pt}
\centering
\caption{Accuracy (\%) of different methods under quantity-based label skews with ViT/B-32 as the backbone. FPL~\cite{menghini2023enhancing} and CPL~\cite{zhang2024candidate} are adopted as baseline pseudo labeling (\textbf{PL}) method. \textbf{Bold} and \underline{underline} represent the best and second-best results.}
\label{tab: accuracy Quantity}
\resizebox{0.8\textwidth}{!}{ 
\begin{tabular}{lcccccccc}
\toprule
{Method} & \textbf{PL} & {{DTD}} & {{RESISC45}} & {{CUB}} & {{UCF101}} & {{CIFAR10}} & {{CIFAR100}} & {Average} \\ \midrule
CLIP & - & {43.24} & {54.51} & \underline{51.28} & {61.00} & {86.93} & {64.17} & 60.19 \\ 
PromptFL~\cite{guo2023promptfl} & FPL & 43.53 & 57.53 & 46.66 & 62.35 & 88.29 & 63.94 & 60.38 \\ 
PromptFL~\cite{guo2023promptfl} & CPL & 43.82 & 61.11 & 47.31 & 64.23 & 88.70 & \underline{67.23} & 62.07 \\
Promptprox~\cite{li2020federated} & FPL & 44.89 & 56.71 & 48.74 & 61.80 & 85.69 & 64.01 & 60.31 \\
Promptprox~\cite{li2020federated} & CPL & 45.53 & 60.77 & 47.61 & 63.86 & {89.21} & 66.49 & 62.25 \\
pFedPrompt~\cite{guo2023pfedprompt} & FPL & 45.09 & 60.46 & 47.71 & 65.39 & 87.54 & 62.82 & 61.50\\
pFedPrompt~\cite{guo2023pfedprompt} & CPL & 45.14 & \underline{66.68} & 50.45 & \underline{65.57} & 87.98 & 65.45 & \underline{63.55} \\
FedOPT~\cite{li2024global} & FPL & \underline{47.65} & 56.04 & 46.61 & 62.95 & \underline{90.74} & 66.76 & 61.79 \\
FedOPT~\cite{li2024global} & CPL & {40.24} & 47.09 & 44.00 & 57.24 & 89.21 & 57.98 & 55.96 \\\cmidrule{1-9} \rowcolor[gray]{0.9}
\textbf{FedCoPL} & \textbf{CoPL} & \textbf{56.18} & \textbf{81.06} & \textbf{56.31} & \textbf{72.03} & \textbf{94.86} & \textbf{73.39}  & \textbf{72.31} \\
\bottomrule
\end{tabular}
}
\end{table*}

\section{Experiments}
\label{sec: experiment}

\subsection{Setups}
\label{Setups}
\textbf{Datasets.}
\label{Datasets}
We evaluated the performance of our method on six public benchmark datasets characterized by varying types of label skew.
Following previous research~\cite{guo2023pfedprompt, li2024global,cui2024harmonizing}, we employed four representative visual classification datasets: DTD~\cite{cimpoi2014describing}, RESISC45~\cite{cheng2017remote}, UCF101~\cite{soomro2012ucf101}, and CUB~\cite{wah2011caltech}, along with two standard federated classification benchmark datasets: CIFAR10 and CIFAR100~\cite{krizhevsky2009learning}.
We partitioned each dataset into distinct training and test sets and further split them into non-overlapping subsets for different clients.
Specifically, we followed the settings outlined in~\cite{li2022federated} and employed two prevalent forms of label skew: quantity-based and Dirichlet-based label skews.
In the quantity-based label skew, all training data is first grouped by label and then partitioned into shards of varying sizes.
In the Dirichlet-based label skew, clients receive samples for each class based on the Dirichlet distribution~\cite{zhu2021federated}.
Here, the parameter $\beta$ controls the degree of label skew, with lower values indicating severe label skew.

\begin{table*}[t]
\setlength{\tabcolsep}{8pt}
\centering
\caption{Accuracies (\%) of combining our pseudo labeling strategy CoPL with existing federated training methods under Dirichlet-based label skews ($\beta=0.1$). FPL~\cite{menghini2023enhancing} is adopted as baseline pseudo labeling (\textbf{PL}) method. (The values) represent the performance gains.}
\label{tab: accuracy our label}
\resizebox{.8\textwidth}{!}{ 
\begin{tabular}{lcllllll}
\toprule
{Method} & \textbf{PL} & {{DTD}} & {{RESISC45}} & {{CUB}} & {{UCF101}} & {{CIFAR10}}    \\ \midrule
PromptFL~\cite{guo2023promptfl} & FPL & 45.79 & 59.76 & 47.29 & 64.39 & 87.51  \\
PromptFL~\cite{guo2023promptfl} & \textbf{CoPL} & 47.34 ({\color{red}+1.55}) & 61.68 ({\color{red}+1.92}) & 49.08 ({\color{red}+1.79}) & 64.97 ({\color{red}+0.58}) & 90.20 ({\color{red}+2.69}) \\
pFedPrompt~\cite{guo2023pfedprompt} & FPL & 44.56 & 65.95 & 48.48 & 64.37 & 87.45 \\
pFedPrompt~\cite{guo2023pfedprompt} & \textbf{CoPL} & 49.25 ({\color{red}+4.69}) & 67.97 ({\color{red}+2.02}) & 52.77 ({\color{red}+2.29}) & 67.93 ({\color{red}+3.56}) & 91.13 ({\color{red}+3.68}) \\
FedOPT~\cite{li2024global} & FPL & 51.97 & 57.76 & 47.99 & 61.83  & 91.32 \\
FedOPT~\cite{li2024global} & \textbf{CoPL} & 57.93 ({\color{red}+5.96}) & 74.85 ({\color{red}+17.09}) & 55.72 ({\color{red}+7.73}) & 69.21 ({\color{red}+7.38}) & 94.02 ({\color{red}+2.70}) \\
\bottomrule
\end{tabular}
}
\end{table*}

\begin{figure*}[t]
  \centering
  \includegraphics[width=\textwidth]{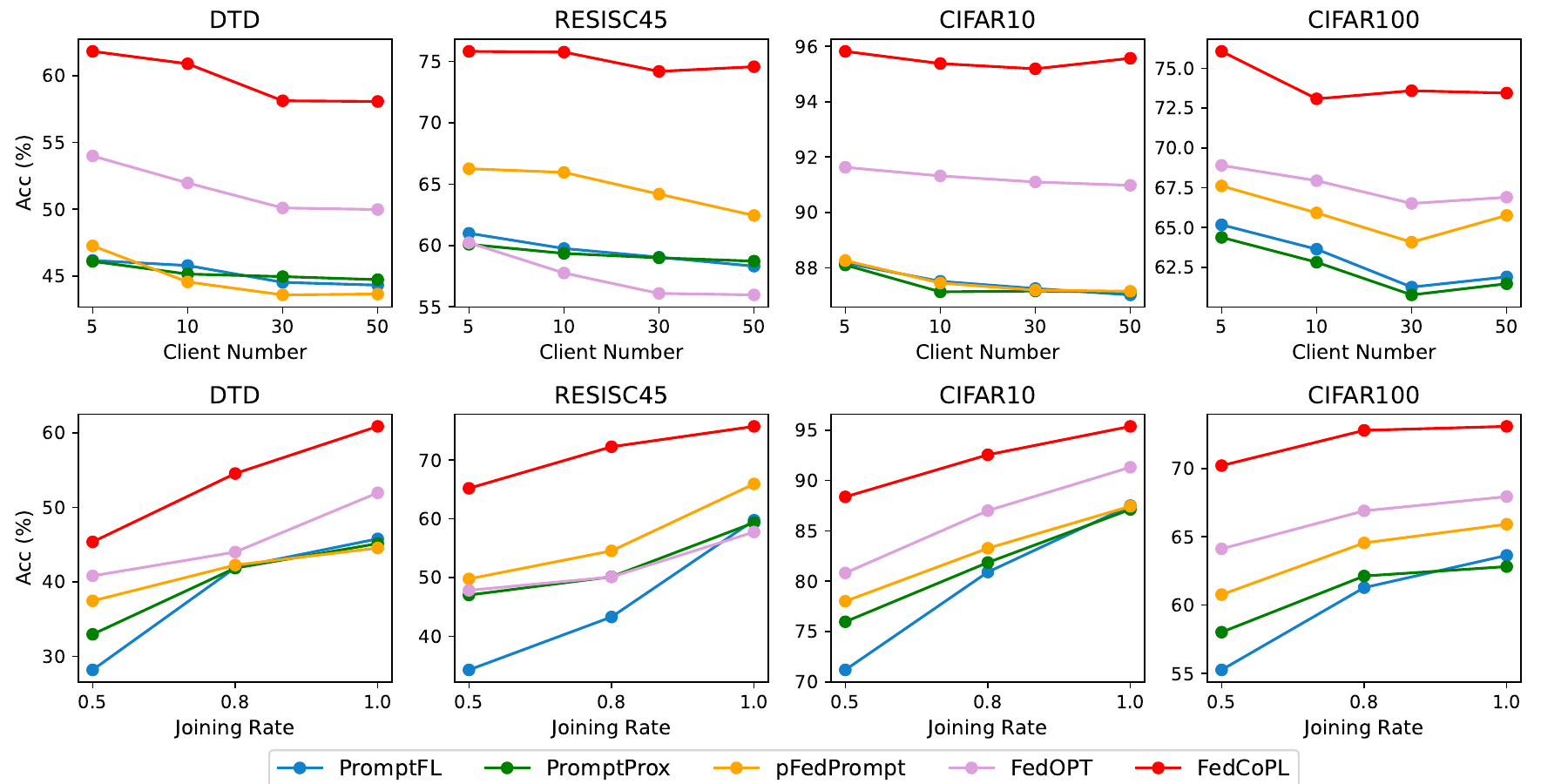}
  \caption{Results of experiments with various client numbers and different client joining rates under Dirichlet-based label skews ($\beta=0.1$). FPL~\cite{menghini2023enhancing} is adopted as the baseline pseudo labeling method.}
  \label{fig: client}
\end{figure*}

\textbf{Baseline methods.}
In our experiments, we compare FedCoPL, with two popular pseudo label selection methods in central unsupervised learning and four supervised federated learning methods. 
Regarding pseudo labeling strategy, \textbf{FPL}~\cite{menghini2023enhancing} selects the most reliable samples based on confidence for each class, while \textbf{CPL}~\cite{zhang2024candidate} generates multiple pseudo labels for each sample through selection at both the sample and category levels. 
Zero-shot prediction involves utilizing the pre-trained CLIP model with a hand-crafted textual prompt template, such as “a photo of a [CLASS],” to predict the test data. 
For federated training methods, \textbf{PromptFL}~\cite{guo2023promptfl} employs shared prompts learned across clients via FedAvg~\cite{mcmahan2017communication}. \textbf{Promptprox}, introduced in~\cite{guo2023promptfl}, is derived from the traditional federated learning technique FedProx~\cite{li2020federated}, which employs the proximal term to constrain the update direction of the local model.
\textbf{pFedPrompt}~\cite{guo2023pfedprompt} learns a unified textual prompt with personalized attention modules for local visual embeddings. 
Finally, \textbf{FedOPT}~\cite{li2024global} performs conditional optimal transport between the global and local textual prompts to effectively fuse global and local knowledge.

\textbf{Implementation details.} 
\label{Implementation details}
All results in this paper are based on a frozen CLIP using two backbones, ViT-B/16~\cite{alexey2020image} and ViT-B32~\cite{alexey2020image}, with ViT-B/32 as the default unless otherwise specified.
We set the number of clients, $K$, to 5 for the CUB~\cite{wah2011caltech} and UCF101~\cite{soomro2012ucf101} datasets, and to 10 for the other datasets, with full client participation as the default setting.
We conduct 20 communication rounds for all experimental datasets, where pseudo labels are updated every 5 rounds using the latest local model. 
Within each communication round, each client performs 10 epochs of local training.
We optimize the prompts using mini-batch Stochastic Gradient Descent (SGD) with a learning rate of 0.1, a momentum of 0.9, and decay following the cosine annealing rule.
We conduct three trials for each experimental setting and report the mean accuracy. 
All experiments are conducted using PyTorch~\cite{paszke2019pytorch} on NVIDIA 3090 GPUs.
More details on the implementation can be found in the supplementary material.

\subsection{Experimental Results}
\label{Experimental Results}
\textbf{Results under Dirichlet-based label skews with various datasets.} 
Table~\ref{tab: accuracy Dirichlet} reports the performance of different methods under Dirichlet-based label skews ($\beta \in {0.1, 0.05}$). 
Our method significantly outperforms state-of-the-art algorithms across all datasets, demonstrating the effectiveness of our cooperative pseudo labeling strategy and partial prompt aggregation protocol.
Notably, FedOPT~\cite{li2024global}, the latest personalized supervised federated prompt tuning method, shows that integrating it with baseline pseudo label selection methods leads to worse performance than zero-shot learning. 
This is attributed to the low accuracy of the pseudo labels chosen by each client, which fail to capture the true local data distribution, and this issue is further exacerbated by personalized training.
A more detailed discussion is provided in the following sections.
In the CUB~\cite{wah2011caltech} dataset, all baseline methods perform worse than zero-shot prediction owing to the worse representativeness of selected pseudo labels.
In contrast, our method consistently achieves superior performance on all datasets, underscoring the importance of the cooperative pseudo labeling strategy and the partial prompt aggregation protocol.

\textbf{Results under quantity-based label skews with various datasets.}
We present the performance of all methods under quantity-based label distribution skews in Table~\ref{tab: accuracy Quantity}, where the parameter is set as $s = C \times 0.2$ across all datasets, with $C$ representing the number of classes in each dataset. 
In this setting, each client possesses samples from only a few classes, which poses challenges for pseudo labeling and model training. 
Owing to this, many baseline results are outperformed by the zero-shot approach. 
For example, FedOPT with CPL performs 7.42\%  and 7.28\% worse than the zero-shot accuracy on the RESISC45 and CUB datasets, respectively .
In contrast, our method maintains strong performance, similar to that observed with Dirichlet-based distributions, further highlighting the effectiveness of FedCoPL.

\subsection{Analysis}
\textbf{Comparation of different pseudo labeling methods.} 
We compare our pseudo labeling method with two baseline methods, FPL~\cite{menghini2023enhancing} and CPL~\cite{zhang2024candidate}. 
As shown in Table~\ref{tab: accuracy Dirichlet} and Table~\ref{tab: accuracy Quantity}, for the same federated training method, CPL outperforms FPL on datasets such as DTD~\cite{cimpoi2014describing} and CUB~\cite{wah2011caltech}, whereas the opposite is true for the CIFAR datasets.
In contrast, our proposed cooperative pseudo labeling strategy consistently outperforms both baseline methods across all datasets.
To further demonstrate the effectiveness of our approach, we integrate our pseudo labeling method into various state-of-the-art supervised federated training frameworks.
As shown in Table~\ref{tab: accuracy our label}, our method significantly enhances the performance of baseline pseudo label selection methods across different datasets, with a notable increase of up to 17.09\% on the RESISC45 dataset. 
The results clearly validate the effectiveness of our pseudo labeling method compared to baseline methods.
\begin{table}[t]
\centering
\caption{Experimental results (\%) using CLIP ViT-B/16 as backbone under Dirichlet-based label skew ($\beta=0.1$). FPL~\cite{menghini2023enhancing} is adopted as the baseline pseudo labeling (\textbf{PL}) method.}
\label{tab: accuracy backbone}
\resizebox{0.45\textwidth}{!}{ 
\begin{tabular}{lccccc}
\toprule
{Method} & \textbf{PL} & {{DTD}} & {{RESISC45}} & {{CUB}} & {{UCF101}}  \\ \midrule
Zero-shot CLIP & - & 42.87 & 56.61 & 55.16 & 65.13 \\
PromptFL~\cite{guo2023promptfl} & FPL & 44.36 & 61.36 & 51.95 & 64.60 \\
PromptProx~\cite{li2020federated} & FPL & 47.12 & 61.01 & 49.44 & 65.31 \\
pFedPrompt~\cite{guo2023pfedprompt} & FPL & 47.65 & 63.18 & 50.19 & 65.19 \\
FedOPT~\cite{li2024global} & FPL  & 46.52 & 58.94 & 49.67 & 65.68  \\ \rowcolor[gray]{0.9}
\textbf{FedCoPL} & \textbf{CoPL} & \textbf{56.83} & \textbf{77.50} & \textbf{59.78} & \textbf{78.29} \\
\bottomrule
\end{tabular}
}
\end{table}

\textbf{Results under different client numbers.}
We analyze the performance of our proposed method compared to baseline methods under varying clients numbers. 
Unless stated otherwise, our experiments are conducted under Dirichlet-based skews with $\beta=0.1$. 
We partition the DTD, RESISC45 and CIFAR datasets among 5, 10, 30, and 50 clients and report their final accuracy in the first row of Figure~\ref{fig: client}. 
Notably, as the number of clients increases, the performance of baseline methods declines significantly, whereas our method consistently maintains high accuracy.
This underscores the robustness of our approach, demonstrating its capability to maintain effective learning performance even under varying numbers of participating clients.

\textbf{Impact of client joining rates.}
In this analysis, we investigate the impact of variations in client participation rates, considering values from ${0.5, 0.8, 1.0}$. 
As illustrated in the second row of Figure~\ref{fig: client}, our approach consistently outperforms competing methods across all participation rates.
As the client participation rate decreases, the performance of all methods deteriorates significantly. 
This instability is anticipated, as a lower participation rate exacerbates the divergence between randomly selected clients and the global model, resulting in unstable convergence.
Nevertheless, our approach maintains superior performance, demonstrating its robustness to fluctuations in client participation rates.

\textbf{Results under different image encoder backbone.}
We further conduct experiments to evaluate the effect of different image encoders.
The comparison results using ViT-B/16 are presented in Table~\ref{tab: accuracy backbone}.
Our approach consistently outperforms previous methods, highlighting the effectiveness of our strategy in improving the performance of smaller image encoders. 
Additional results using RN50 as the image encoder backbone can be found in the supplementary material.
These experiments emphasize the robustness of FedCoPL in real-world FL scenarios, particularly when clients have limited computational resources.

\textbf{Effectiveness of each component.}
Our approach consists of two key modules: a cooperative pseudo labeling strategy and a partial prompt aggregation protocol. 
The results presented in Table~\ref{tab: accuracy loss} demonstrate the effectiveness of the two filtering criteria and global allocation.
To further validate the effectiveness of our partial prompt aggregation, we conduct an additional experiment, as shown in Table~\ref{tab: ablation_partial}. 
Compared to aggregating two modal prompts, aggregating textual prompts only, or retaining two modal prompts locally, the partial prompt aggregation yields the optimal results, demonstrating its effectiveness.

\begin{table}[t]
\centering
\caption{Ablation study. Accuracies (\%) under Dirichlet-based label skews. Conf. and Ent. denote confidence-based and entropy-based filters. G.A. is the global allocation of pseudo labels.}
\label{tab: accuracy loss}
\resizebox{0.45\textwidth}{!}{ 
\begin{tabular}{ccccccc}
\toprule
 \textbf{Conf.} & \textbf{Ent.} & \textbf{G.A.}  & {{DTD}} & {{RESISC45}} & {{CUB}} & {{UCF101}}  \\ \midrule
- & - & - &  45.79 & 59.76 & 47.29 & 64.39 \\
- & - & \cmark &  46.35 & 69.60 & 51.98 & 65.31 \\
- & \cmark & \cmark &  58.83 & 72.28 & 54.91 & 70.23 \\
\cmark & - & \cmark &  55.59 & 73.66 & 50.38 & 69.13 \\
\cmark & \cmark & \cmark &  60.89 & 75.76 & 56.09 & 73.20 \\ 
\bottomrule
\end{tabular}
}
\end{table}

\begin{table}[t]
\centering
\setlength{\abovecaptionskip}{0.1cm}
\caption{Ablation results (\%) comparing the effects of aggregating textual prompts (\textbf{T.P.}) only, visual prompts (\textbf{V.P.}) only, both, or none under Dirichlet-based label skew ($\beta = 0.1$). }
\label{tab: ablation_partial}{}
\resizebox{.43\textwidth}{!}{ 
\begin{tabular}{cccccc}
\toprule
\textbf{T.P.} & \textbf{V.P.} & DTD & RESISC45 & CUB & UCF101\\ \midrule
- & - & 50.45 & 61.72 & 49.08 & 66.02 \\
\cmark & \cmark & 47.34 & 61.68 & 49.08 & 64.97 \\
\cmark & -  & 55.31 & 67.12 & 51.76 & 69.53 \\
- & \cmark & 60.89 & 75.76 & 56.09 & 73.20 \\
\bottomrule
\end{tabular}
}
\end{table}

\section{Conclusion}
\label{5_conclusion}
In this paper, we extend unsupervised federated learning to classification tasks using CLIP for the first time.
In this setting, clients with pre-trained CLIP models and unlabeled data collaborate in training to enhance performance without data sharing.
To address the challenge of CLIP's internet bias and potential label skew among clients, we propose FedCoPL, which includes a cooperative pseudo labeling strategy and a partial prompt aggregation protocol.
Specifically, each client estimates and uploads its pseudo label distribution, and the server adjusts and redistributes them to mitigate global imbalance across categories.
Moreover, the aggregation protocol aggregates only visual prompts on the server to improve global performance, while textual prompts are kept locally for better personalization by each client.
Extensive experimental results demonstrate that our FedCoPL outperforms baseline methods across various degrees of label skew.
In future work, we will conduct a theoretical analysis of FedCoPL, including convergence, privacy, fairness, and other pertinent considerations. 

\section*{Acknowledgments}
This work was funded by the National Natural Science Foundation of China under Grants (62276256, U2441251), the Young Elite Scientists Sponsorship Program by CAST (2023QNRC001), and the Young Scientists Fund of the State Key Laboratory of Multimodal Artificial Intelligence Systems (ES2P100117).

{
    \small
    \bibliographystyle{ieeenat_fullname}
    \bibliography{main}
}

\clearpage
\setcounter{page}{1}
\maketitlesupplementary
\vspace{1em}

\section{Algorithm Flow of FedCoPL}
\label{The Pseudocode of Our Method}
We summarize the procedure of FedCoPL in Algorithm~\ref{alg:FedPP}.
\begin{algorithm}[h]
    \renewcommand{\algorithmicrequire}{\textbf{Input:}}
    \renewcommand{\algorithmicensure}{\textbf{Output:}}
        \caption{\textbf{FedCoPL}}
    \label{alg:FedPP}
    \begin{algorithmic}[1]
        \REQUIRE number of communication round $T$, client number $K$, unlabeled dataset $\{D_k\}^K_{k=1}$, client participating rate $R$, number of local update epochs $E$, batch size $B$, learning rate $\eta$, pseudo labels update interval $Q$.
        \ENSURE the global visual prompt $P^v$ and personalized textual prompts $\{P^t_k\}^K_{k=1}$.
        \STATE Initialize $P^{v}$ , $\{P^{t}_k\}^K_{k=1}$
        \STATE $m \gets \max( \lfloor R \cdot K \rfloor, 1)$
        \FOR{communication round $r=1,2,\cdots,T$}
        
            \IF{r \% Q = 0} 
            \STATE  \textit{\textcolor{gray}{\# cooperative pseudo labeling}}
            \FOR {$k=1, ..., K$}
            \STATE  Obtain the estimated set $D^{est}_k$ with Eq. (2). \\            
            \STATE Obtain the estimated statistics $\widetilde{U}_k$ with Eq. (3). 
            \STATE  Obtain $\widetilde{D}_k$ by selecting the most confident samples according to the capacity $\widetilde{U}_{k}$.
            \ENDFOR
            \ENDIF
            \STATE $M \gets $ Randomly select a subset containing $m$ clients.
            
            \STATE \textit{\textcolor{gray}{\# local update}} 
            \FOR{each client $k \in M$}
                \STATE Initialize local visual prompt $P^{v}_k \leftarrow P^{v}$
                \vspace{0.1cm}
                \FOR{\textit{each batch} $\mathcal{B}_{i} = \{\textit{x},\hat{\textit{y}}\} \in \widetilde{D}_k$} \vspace{0.4\baselineskip}   
                    \STATE $P^{v}_k \leftarrow P^{v}_k - \eta \nabla {\mathcal{L}({P^{v}_k;\mathcal{B}_i)}}$ 
                    \vspace{0.3\baselineskip}
                    \STATE $P^{t}_k \leftarrow P^{t}_k - \eta \nabla {\mathcal{L}({P^{t}_k;\mathcal{B}_i)}}$ \textcolor{gray}{\# $\mathcal{L}$ \textit{is cross-entropy loss}}
                \ENDFOR
            \ENDFOR
            \STATE Obtain aggregated visual prompt $P^{v}$ with Eq. (4).            
        \ENDFOR 
\end{algorithmic}
\end{algorithm}

\section{Drift Diversity}
Following~\cite{li2023effectiveness}, we employ drift diversity to assess magnitude differences, which is defined as follows: 
\begin{equation}
\xi^r := \frac{\sum_{k=1}^{K} \| m_k^r \|^2}{\| \sum_{k=1}^{K} m_k^r \|^2}
\quad \text{with} \quad
m_k^r = P_{k}^r - P^{r-1}
\end{equation}
where $P_{k}^r$ is updated prompt of client $k$ in round $r$ and $P^{r-1}$ is aggregated prompt on the server in round $r-1$.
\begin{figure}
    \small
    \setlength\tabcolsep{1mm}
    \renewcommand\arraystretch{0.1}
    \begin{tabular}{cc}
        \includegraphics[width=0.44\linewidth]{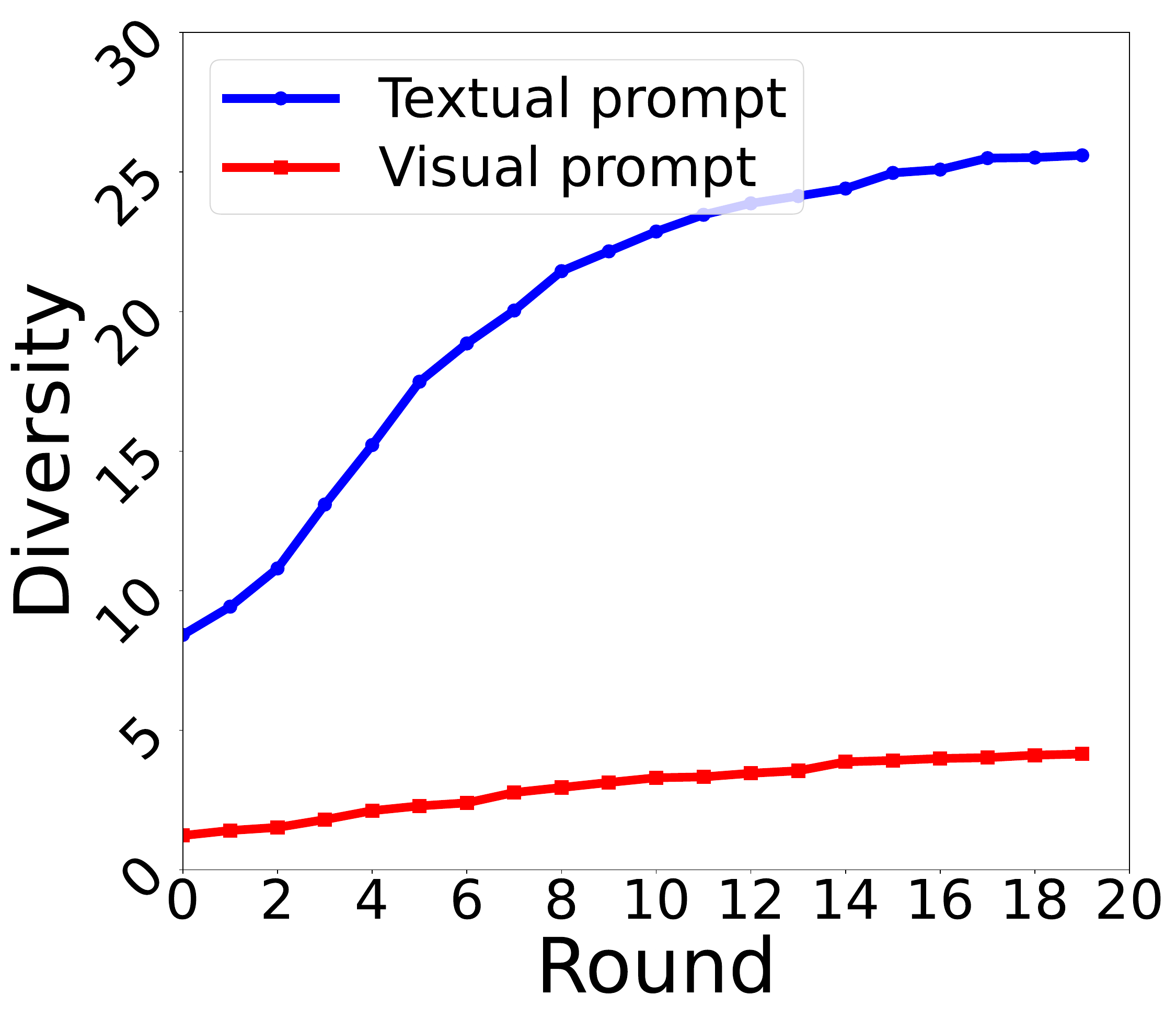} &		
        \includegraphics[width=0.44\linewidth, clip]{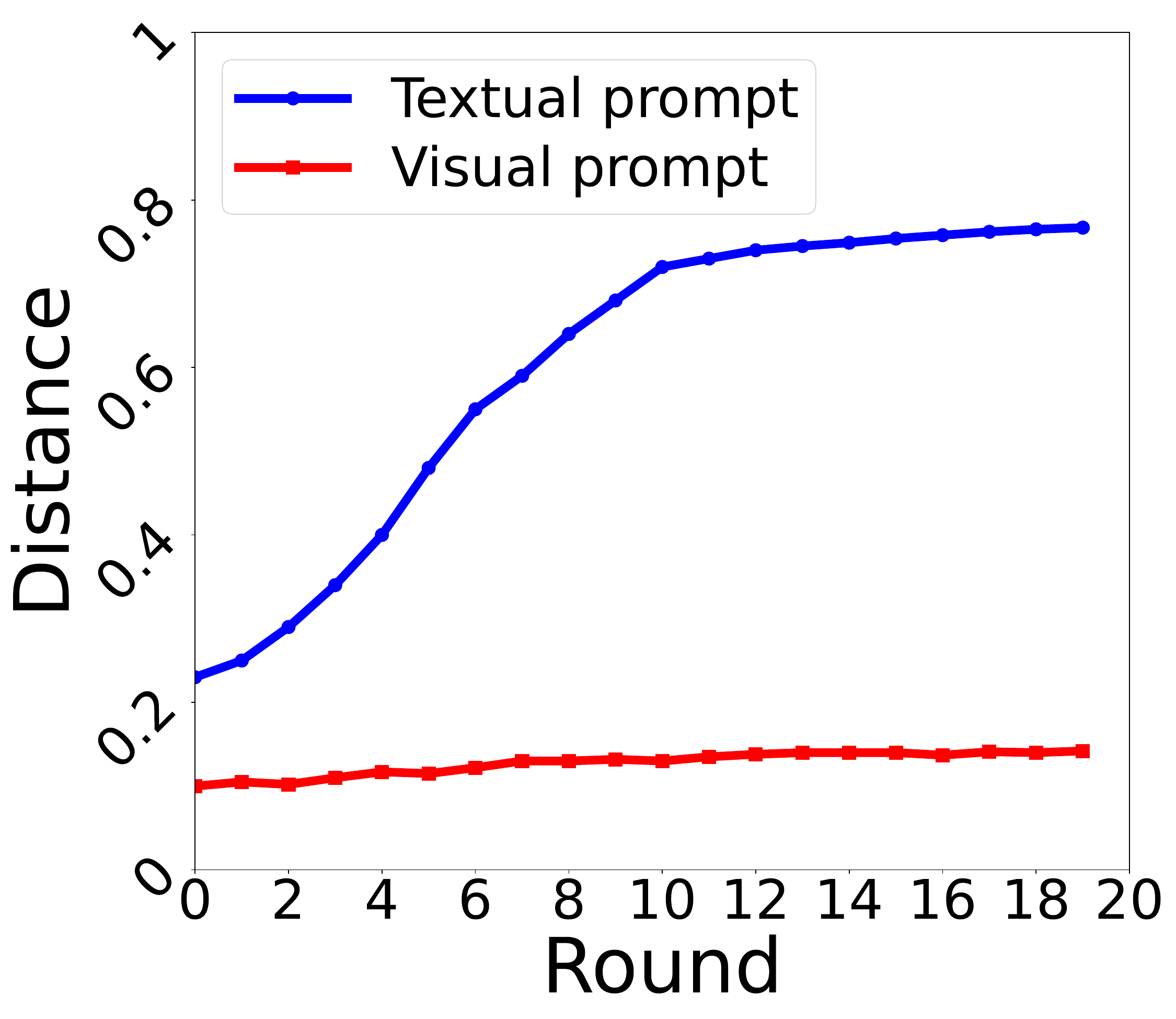} 
    \\[1.0mm]
        (a) Drift Diversity. & (b) Cosine Distance.
    \end{tabular}
    \caption{Drift diversity and cosine distance of prompts among clients during training in DTD~\cite{cimpoi2014describing} dataset.
    The differences observed in textual prompts are significantly greater than those found in visual prompts.}
    \label{fig: shift_DTD}
\end{figure}

\begin{figure}
    \small
    \setlength\tabcolsep{1mm}
    \renewcommand\arraystretch{0.1}
    \begin{tabular}{cc}
        \includegraphics[width=0.44\linewidth]{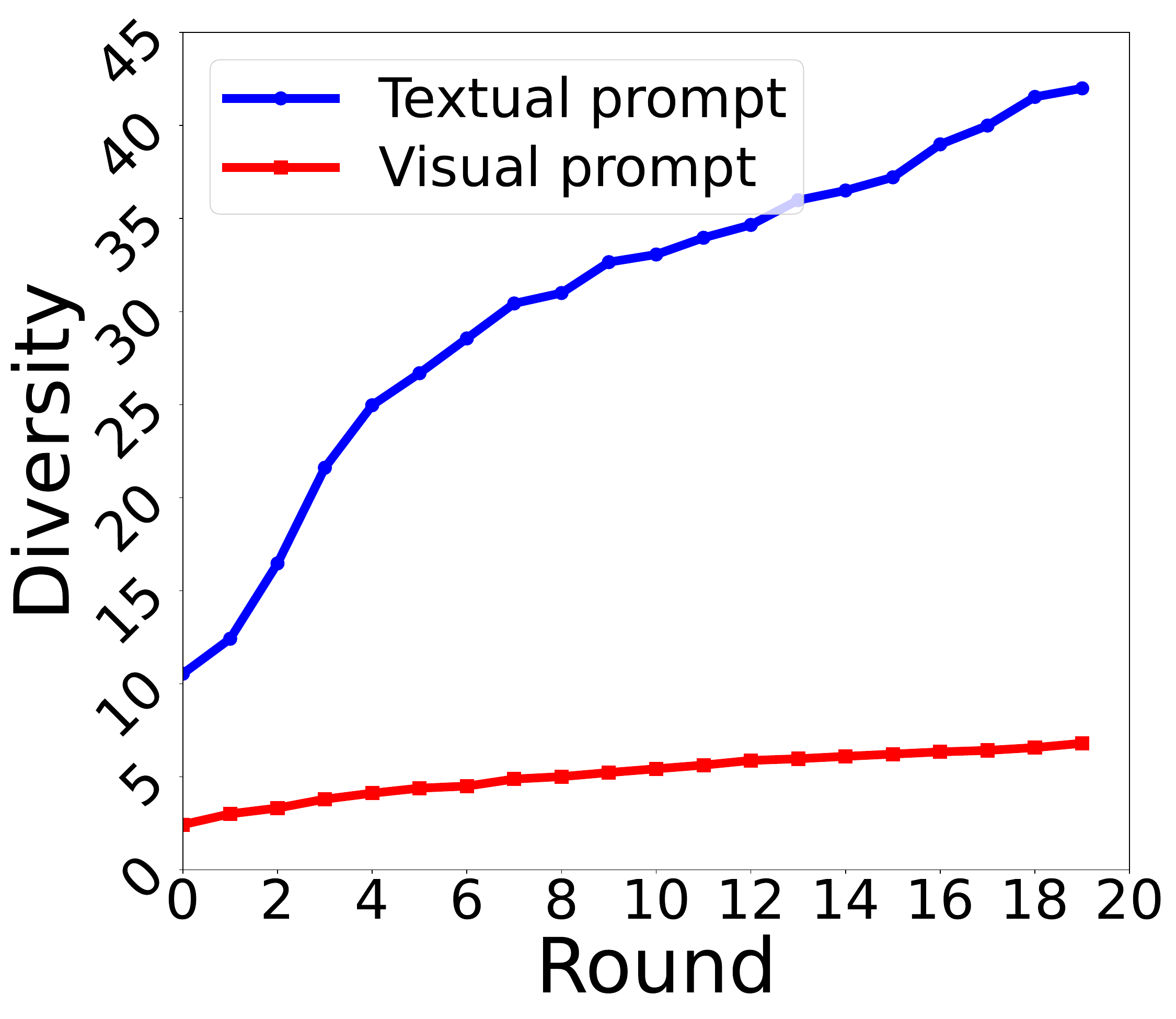} &		
        \includegraphics[width=0.44\linewidth, clip]{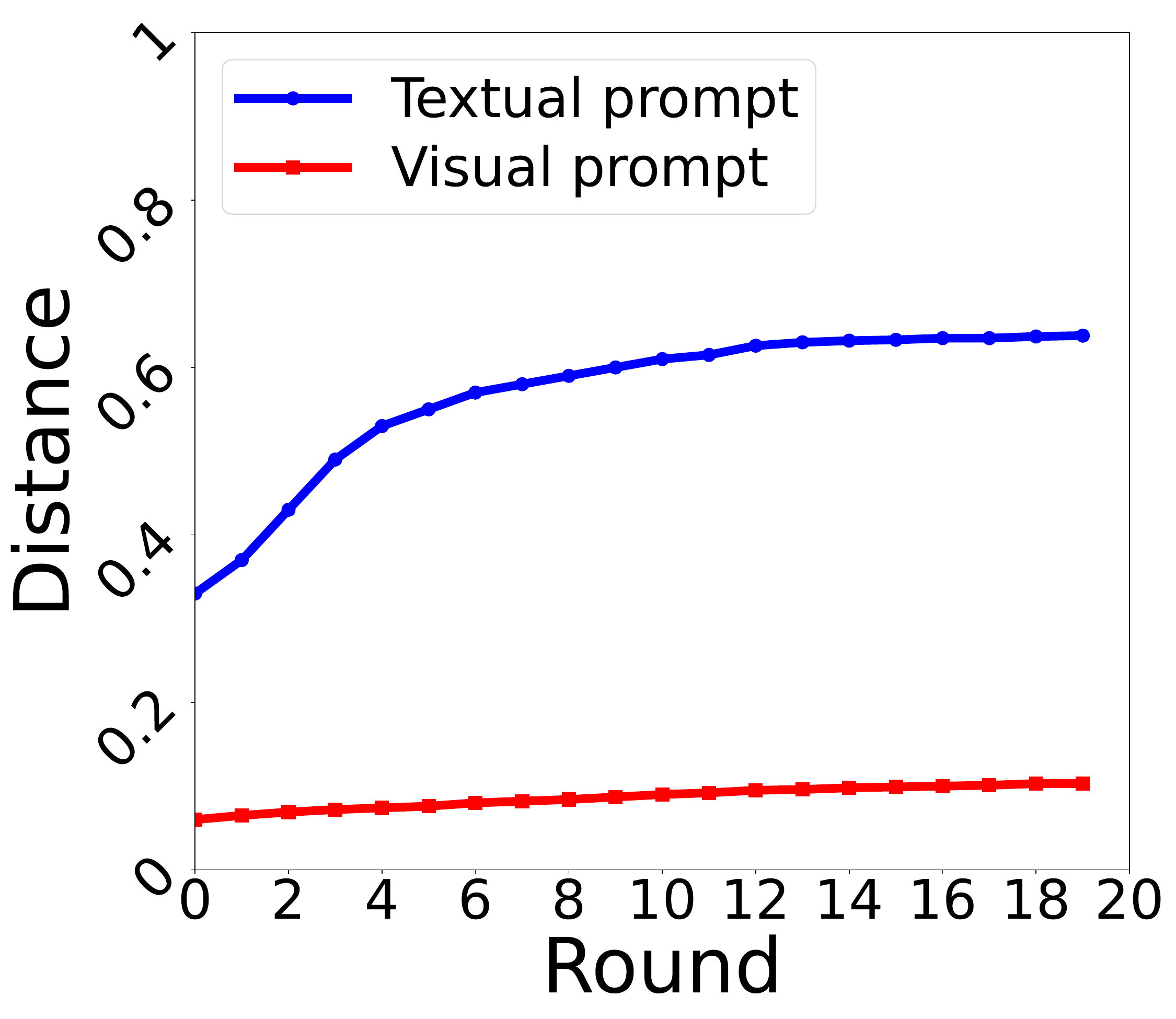} 
    \\[1.0mm]
        (a) Drift Diversity. & (b) Cosine Distance.
    \end{tabular}
    \caption{Drift diversity and cosine distance of prompts among clients during training in RESISC45~\cite{cheng2017remote} dataset.}
    \label{fig: shift_RESISC45}
\end{figure}

Besides, we measure the differences in both textual and visual prompts across all clients using drift diversity~\cite{li2023effectiveness} and cosine distance in RESISC45 and DTD datasets,  which respectively reflect the diversity in the amount and direction of prompts updates among clients, as shown in Figure~\ref{fig: shift_DTD} and Figure~\ref{fig: shift_RESISC45}.
These results prove that the differences in textual prompts are significantly greater than those in visual prompts, which confirms our hypothesis that visual prompts tend to be more similar, while textual prompts exhibit greater variability.

\begin{figure*}[t]
		\small
		\setlength\tabcolsep{1mm}
		\renewcommand\arraystretch{0.1}
		\begin{tabular}{ccc}
            \includegraphics[width=0.3\linewidth]{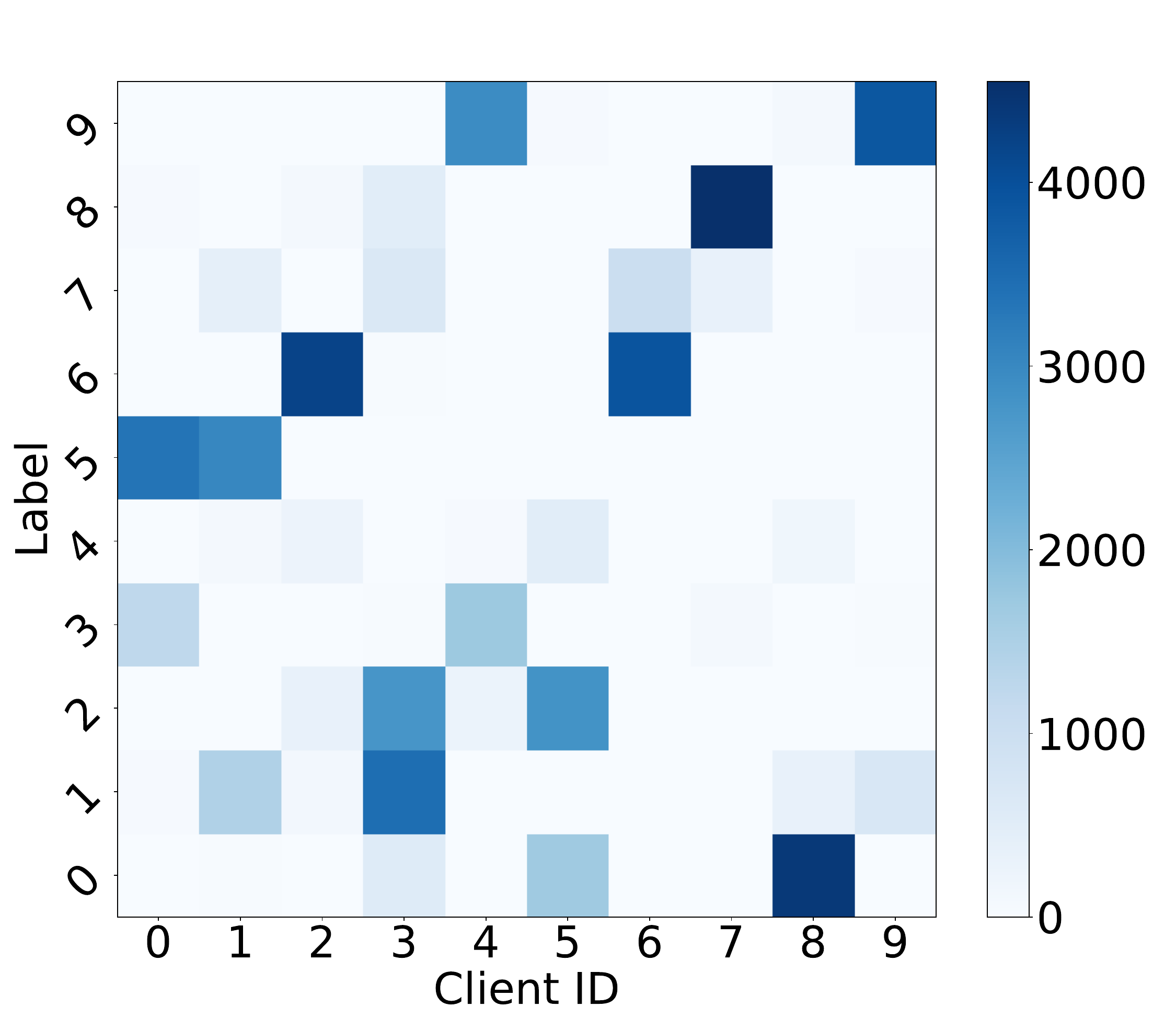} &		
            \includegraphics[width=0.3\linewidth, clip]{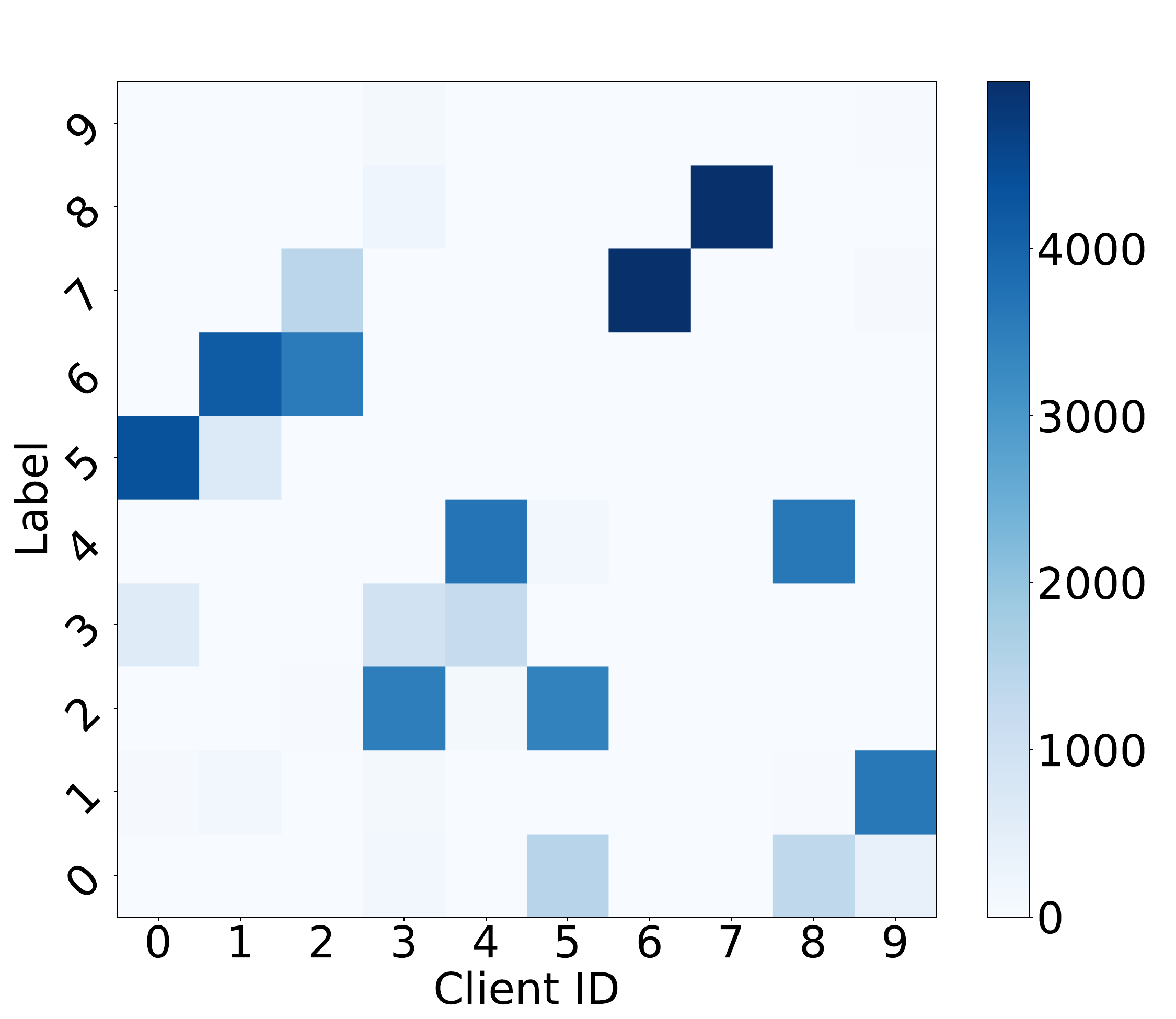} &
            \includegraphics[width=0.3\linewidth, clip]{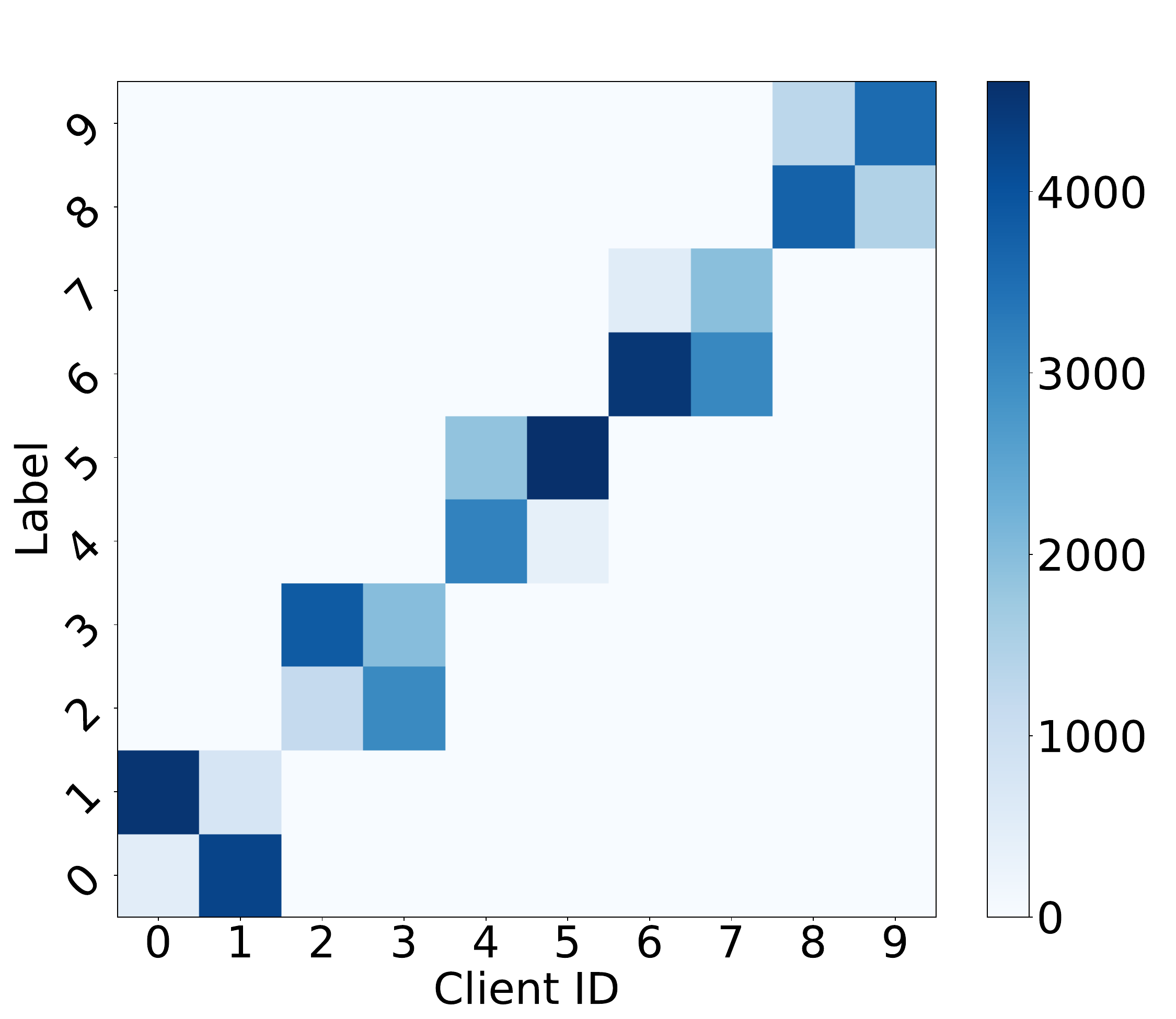}
			\\[1.0mm]
			(a) Dirichlet-based ($\beta =0.1$). & (b) Dirichlet-based ($\beta =0.05$). & (c) Quantity-based.
		\end{tabular}
		\caption{(a) and (b) depict label skews of Dirichlet-based label skews and (c) presents the quantity-based label skew.}
		\label{fig: all distribution}
\end{figure*}

\section{More Experimental Details}
\label{Details of Dataset Setup.}
\textbf{Dataset setup.}
We evaluate our approach on six diverse visual classification datasets. 
Table~\ref{tab:dataset} summarizes the key statistics of each dataset, including the original task domain, number of classes, and the number of training and testing samples.
\begin{table}[ht]
\centering
\caption{The detailed statistics of datasets used in experiments.}
\resizebox{0.45\textwidth}{!}{ 
\begin{tabular}{lcccc}
\toprule
{Dataset} & {Task}  & {Classes} & {Training Size} & {Testing Size} \\ \midrule
CUB & Image classification  & 200    & 5,594  & 5,794  \\ 
RESISC45 & Scene classification & 45 & 6,300 & 25,200  \\ 
UCF101 & Action recognition & 101 & 7,639 & 3,783  \\ 
DTD  & Texture recognition & 47   & 2,820  & 1,692  \\ 
CIFAR10 & Image classification  & 10 & 50,000  & 10,000 \\ 
CIFAR100 & Image classification  & 100  & 50,000& 10,000  \\ \bottomrule
\end{tabular}
}
\label{tab:dataset}
\end{table}
We simulate data heterogeneity using both quantity-based and Dirichlet-based label skews. 
In the quantity-based setting, each client is assigned a fixed number of classes: 10 for DTD, 58 for RESISC45, 66 for CUB, 30 for UCF101, 2 for CIFAR-10, and 20 for CIFAR-100. 
For the Dirichlet-based label skew, we generate client data using Dirichlet distributions with concentration parameters $\beta = \{0.1, 0.05\}$. 
To illustrate the label distribution under each setting, we visualize the client-level class allocations on CIFAR-10, as shown in Fig.~\ref{fig: all distribution}.

\textbf{Implementation details.}
All input images are resized to $224 \times 224$ and partitioned into $14 \times 14$ patches with an embedding dimension of 768. 
We incorporate deep visual prompts by appending trainable prompts of size $5 \times 867$ to the output of each transformer layer in the visual encoder.
For the text encoder, we use prompts of length 16 with a dimensionality of 512.
The batch size is set to 64 for both training and evaluation.

\section{More Experiments Results}
\label{additional experiments results}
\textbf{Results under different image encoder backbones.}
We further conduct experiments to evaluate the impact of different image encoders on model performance.
The comparison results using RN50 are summarized in Table~\ref{tab: accuracy backbone RN50}.
These results reveal that as the zero-shot performance of the pre-trained image encoder declines, the performance of all methods deteriorates sharply.
Notably, on the CUB dataset, all baseline methods achieve lower accuracy than the zero-shot baseline, whereas our proposed method surpasses the zero-shot accuracy.
Overall, our approach consistently outperforms prior methods, underscoring the effectiveness of our strategy in enhancing the performance of smaller image encoders.
These findings highlight the robustness of FedCoPL in real-world federated learning scenarios, particularly under limited computational resources.
\begin{table}[t]
\centering
\caption{Experiments using CLIP RN50 as base model under Dirichlet-based label skews ($\beta=0.1$) across four datasets. FPL~\cite{menghini2023enhancing} is adopted as the baseline pseudo labeling (\textbf{PL}) method.}
\label{tab: accuracy backbone RN50}
\resizebox{0.45\textwidth}{!}{ 
\begin{tabular}{lccccc}
\toprule
{Method} & \textbf{PL} & {\textbf{DTD}} & {\textbf{RESISC45}}  & {\textbf{UCF101}} & {\textbf{CIFAR10}} \\ \midrule
Zero-shot CLIP & - & 41.62 & 52.12  & 65.13 & 74.80\\
PromptFL~\cite{guo2023promptfl} & FPL & 42.36 & 56.16  & 62.90 & 78.01\\
PromptProx~\cite{li2020federated} & FPL & 43.12 & 56.54  & 64.37 & 78.77\\
pFedPrompt~\cite{guo2023pfedprompt} & FPL & 43.65 & 57.13  & 64.97 & 79.69\\
FedOPT~\cite{li2024global} & FPL  & 45.52 & 58.94  & 65.54 & 80.13 \\ \rowcolor[gray]{0.9}
\textbf{FedCoPL} & \textbf{CoPL} & \textbf{51.83} & \textbf{70.78}  & \textbf{76.48} & \textbf{84.85}\\
\bottomrule
\end{tabular}
}
\end{table}

\begin{table}[h]
\centering
\caption{Pseudo label accuracy (\%) of different methods with Dirichlet-based label skews ($\beta=0.1$, $\beta=0.05$) and quantity-based label skew on various datasets.}
\label{tab: accuracy pseudo_1}
\resizebox{.48\textwidth}{!}{ 
\begin{tabular}{lcccccc}
\toprule
Method & {{DTD}} & {{RESISC45}} & {{CUB}} & {{UCF101}} & {{CIFAR10}} & {{CIFAR100}}  \\ \midrule
 \multicolumn{7}{c}{Dirichlet-based label skew ($\beta=0.1$)} \\ \midrule
FPL & 66.36 & 77.06 & 77.97 & 80.31 & 91.02 & 83.05 \\
CPL  & 69.72 & 78.21 & 80.64 & 81.17 & 92.18 & 84.81 \\ \rowcolor[gray]{0.9}
\textbf{CoPL (Ours)} & \textbf{78.74} & \textbf{85.73} & \textbf{89.30} & \textbf{85.12} & \textbf{96.07} & \textbf{88.13} \\ \midrule

 \multicolumn{7}{c}{Dirichlet-based label skew ($\beta=0.05$)} \\ \midrule
FPL & 63.54 & 72.51 & 76.72 & 78.19 & 90.26 & 81.87 \\
CPL  & 65.45 & 76.68 & 78.24 & 80.71 & 90.37 & 82.08 \\ \rowcolor[gray]{0.9}
\textbf{CoPL (Ours)} & \textbf{75.82} & \textbf{85.09} & \textbf{88.62} & \textbf{83.90} & \textbf{95.26} & \textbf{86.97} \\ \midrule

 \multicolumn{7}{c}{Quantity-based label skew} \\ \midrule
FPL & 55.18 & 61.02 & 62.79 & 65.44 & 87.79 & 76.18 \\
CPL  & 56.50 & 64.90 & 66.92 & 68.66 & 88.50 & 76.91 \\ \rowcolor[gray]{0.9}
\textbf{CoPL (Ours)} & \textbf{68.29} & \textbf{78.87} & \textbf{79.21} & \textbf{79.76} & \textbf{94.24} & \textbf{85.03} \\ \bottomrule

\end{tabular}
}
\end{table}


\textbf{Comparison of pseudo-label accuracy.}
As shown in Table~\ref{tab: accuracy pseudo_1}, we report the accuracy of various pseudo labeling methods based on CLIP’s zero-shot predictions in Dirichlet-based and quantity-based label skews.
The proposed pseudo label selection strategy consistently outperforms baseline approaches across multiple datasets.
These results underscore the effectiveness of the global pseudo label allocation strategy, which provides a robust foundation for subsequent model training.
By explicitly accounting for global class distributions and aggregating client-level pseudo label distributions, our method effectively alleviates label skew across clients and enhances the consistency of assigned pseudo labels.


\begin{table*}[!t]
\centering
\caption{Performance (\%) of FedCoPL under different values of hyperparameter $\tau _1$ with Dirichlet-based label skews ($\beta$ = 0.1) on four datasets ($\tau _2 = 0.50$).}
\label{tab: hyperparameter 1}
\resizebox{.95\textwidth}{!}{ 
\begin{tabular}{lccccccccccccccc}
\toprule
$\tau _1$ & {{0.10}} & {{0.20}} & {{0.30}} & {{0.40}} & {{0.44}} & {{0.46}} & {{0.48}} & {{0.50}} & {{0.52}} & {{0.54}} & {{0.56}} & {{0.60}} & {{0.70}} & {{0.80}} & {{0.90}} \\ \midrule
{DTD} & 55.89  &  55.92  &  56.01  &  56.13 & 58.90 & 59.76 & 60.62 & 60.89 & \textbf{60.92} & 60.53 & 60.02  &  54.85  &  51.09  &  47.56  & 36.16   \\
{RESICS45} & 75.29  &  75.15  &  74.92  &  75.31  & 75.03 & 75.84 & \textbf{76.31} & 75.76 & 75.61 & 74.88 & 74.07  &  71.14  &  61.27  &  54.98  &  47.58  \\
{CIFAR10} &  83.38  &  80.77  &  82.84  &  85.34  & 92.68 & 93.83 & 95.24 & 95.38 & 95.52 & \textbf{95.71} & 94.10  &  90.41  &  77.56  &  49.76  &  55.59  \\
{CIFAR100} &  72.14  &  72.64  &  72.12  &   72.29  & 72.54 & 72.89 & 73.31& \textbf{73.59} & 73.01 & 72.65 & 73.18  &  72.78  &  72.21  &  73.05  & 68.44  \\
\bottomrule
\end{tabular}
}
\end{table*}
\begin{table*}[!t]
\centering
    \caption{Performance (\%) of the FedCoPL under different values of hyperparameter $\tau _2$ with Dirichlet-based label skews ($\beta$ = 0.1) on four datasets ($\tau _1 = 0.50$).}
\label{tab: hyperparameter 2}
\resizebox{.95\textwidth}{!}{ 
\begin{tabular}{lccccccccccccccc}
\toprule
$\tau _2$ & {{0.10}} & {{0.20}} & {{0.30}} & {{0.40}} & {{0.44}} & {{0.46}} & {{0.48}} & {{0.50}} & {{0.52}} & {{0.54}} & {{0.56}}  & {{0.60}} & {{0.70}} & {{0.80}} & {{0.90}} \\ \midrule
{DTD} & 58.04  &  58.07  &  58.38  & 57.98 & 58.10 & 58.92 & 59.93 & \textbf{60.89} & 60.25 & 59.67 & 58.99  &  57.44 &   56.67 &   49.22 & 37.92  \\
{RESICS45} & 71.92  & 75.12   &  75.82  & 75.07 & 74.85 & 75.12 & 75.54 & \textbf{75.76} & 75.04 & 74.37 & 73.89  & 72.43  &  66.25  &  64.49  & 46.80  \\
{CIFAR10}  &  81.01 & 81.92   &  83.87  &  82.73  & 94.08 & 94.93 & 95.22 & 95.38 & \textbf{95.60} & 95.19 & 94.68  &  88.36 &  71.55  &  42.33  & 50.92 \\
{CIFAR100} & 72.33  &  72.39  &  72.35  &  71.20 & 71.33 & 72.48 & 72.70 & 73.59 & \textbf{73.62} & 72.81 & 71.97  & 71.13  &  71.37  &  71.40  & 67.20 \\
\bottomrule
\end{tabular}
}
\end{table*}

\begin{figure*}[!t]
  \centering
   \setlength{\abovecaptionskip}{0.cm}
   \includegraphics[width=0.9\textwidth]{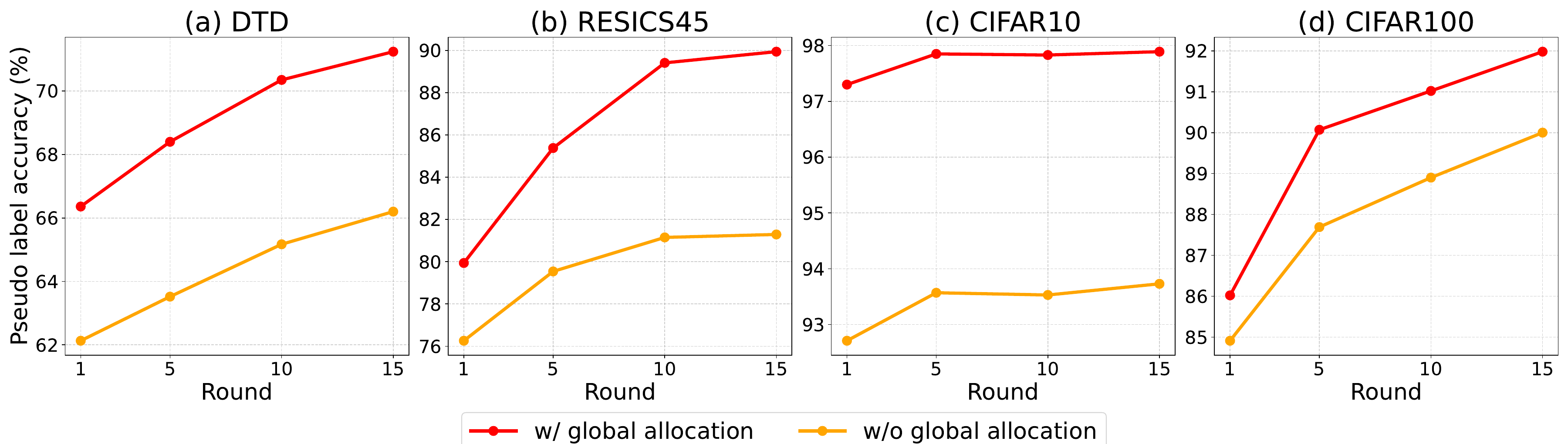}
  \caption{Pseudo label accuracy of our method with and without global pseudo label allocation on four datasets.}
    \label{pseudo_acc}
\end{figure*}

\textbf{Sensitivity analysis of hyperparameters $\tau _1$ and $\tau _2$.}
To demonstrate the robustness of our method with respect to hyperparameter selection, we conduct experiments using a range of values for $\tau_1$ and $\tau_2$ on the DTD, RESISC45, CIFAR10, and CIFAR100 datasets.
The results, summarized in Table~\ref{tab: hyperparameter 1} and Table~\ref{tab: hyperparameter 2}, show that our method exhibits strong insensitivity to the choice of $\tau_1$ and $\tau_2$.
Specifically, across $\tau_1 \in \{0.44, 0.46, 0.48, 0.50, 0.52, 0.54, 0.56\}$ (with $\tau_2 = 0.50$) and $\tau_2 \in \{0.44, 0.46, 0.48, 0.50, 0.52, 0.54, 0.56\}$ (with $\tau_1 = 0.50$), our approach consistently achieves approximately 60\% on DTD, 76\% on RESISC45, 95\% on CIFAR10, and 73\% on CIFAR100.
This stable performance highlights the method’s robustness and its capacity to deliver reliable results across varying hyperparameter settings.
Notably, in our experiments, we did not perform extensive tuning to identify the optimal values of $\tau_1$ and $\tau_2$.
Instead, we simply set both to the default value of 0.5, which may not represent the best possible configuration. 
This further emphasizes the effectiveness of our method, even without fine-grained hyperparameter selection.

\textbf{Ablation study on global allocation of pseudo labels.}
In this subsection, we conduct an ablation study on the global pseudo label allocation strategy to validate its effectiveness. As shown in Fig.~\ref{pseudo_acc}, we present the accuracy of pseudo labels with and without the global pseudo label allocation. The results show that global allocation not only achieves higher pseudo label accuracy but also leads to more stable and consistent convergence during training. This suggests that global pseudo label allocation among clients helps mitigate the influence of label skews, which are common challenges in federated learning. Moreover, improved pseudo label quality highlights the practical benefits of the proposed global allocation strategy.


\end{document}